\documentclass{article}     % uncomment this line when writing your paper

\usepackage{arxiv}
\usepackage[utf8]{inputenc} % allow utf-8 input
\usepackage[T1]{fontenc}    % use 8-bit T1 fonts
\usepackage{hyperref}       % hyperlinks
\usepackage{url}            % simple URL typesetting
\usepackage{booktabs}       % professional-quality tables
\usepackage{amsfonts}       % blackboard math symbols
\usepackage{nicefrac}       % compact symbols for 1/2, etc.
\usepackage{microtype}      % microtypography
\usepackage{lipsum}
\usepackage{graphicx}
\usepackage{caption}
\usepackage{subcaption}
\usepackage{multirow}
\usepackage{mathrsfs}
\usepackage{float}
\usepackage{amssymb}
\usepackage{stackrel}
\usepackage{algorithm}
\usepackage{algpseudocode}
\usepackage{threeparttable}
\usepackage{bm}

\usepackage{amsmath}
\usepackage{array}
\usepackage{textcomp}
\usepackage{stfloats}
\usepackage{verbatim}
\usepackage{balance}

\usepackage{mathrsfs}
\usepackage{float}
\usepackage{amssymb}
\usepackage{stackrel}
\usepackage{algorithm}
\usepackage{algpseudocode}
\usepackage{threeparttable}
\usepackage{bm}
\title{Proprioceptive State Estimation for \\Amphibious Tactile Sensing}
\author{
  Ning Guo\(^{\#}\), Xudong Han\(^{\#}\)\\
  Department of Mechanical and Energy Engineering\\
  Southern University of Science and Technology\\
  Shenzhen, China 518055\\
  \And
  Shuqiao Zhong\(^{\#}\), Zhiyuan Zhou, Jian Lin\\
  Department of Ocean Science and Engineering\\
  Southern University of Science and Technology\\
  Shenzhen, China 518055\\
  \And
  Jiansheng Dai\\
  Institute of Robotics\\
  Southern University of Science and Technology\\
  Shenzhen, China 518055\\
  \And
  Fang Wan\(^{*}\)\\
  School of Design\\
  Southern University of Science and Technology\\
  Shenzhen, China 518055\\
  \texttt{wanf@sustech.edu.cn}\\
  \And
  Chaoyang Song\thanks{Corresponding Authors.}\\
  Design \& Learning Research Group\\
  \texttt{songcy@ieee.org}\\
}
\begin{document}
\maketitle
%%%%%%%%%%%%%%%%%%%%%%%%%%%%%%%%%%%%%%%%%
\begin{abstract}

    This paper presents a novel vision-based proprioception approach for a soft robotic finger that can estimate and reconstruct tactile interactions in both terrestrial and aquatic environments. The key to this system lies in the finger's unique metamaterial structure, which facilitates omni-directional passive adaptation during grasping, protecting delicate objects across diverse scenarios. A compact in-finger camera captures high-framerate images of the finger's deformation during contact, extracting crucial tactile data in real-time. We present a volumetric discretized model of the soft finger and use the geometry constraints captured by the camera to find the optimal estimation of the deformed shape. The approach is benchmarked using a motion capture system with sparse markers and a haptic device with dense measurements. Both results show state-of-the-art accuracies, with a median error of 1.96 mm for overall body deformation, corresponding to 2.1$\%$ of the finger's length. More importantly, the state estimation is robust in both on-land and underwater environments as we demonstrate its usage for underwater object shape sensing. This combination of passive adaptation and real-time tactile sensing paves the way for amphibious robotic grasping applications.
    
\end{abstract}
%%%%%%%%%%%%%%%%%%%%%%%%%%%%%%%%%%%%%%%%%
%%%%%%%%%%%%%%%%%%%%%%%%%%%%%%%%%
\section{Introduction}
\label{sec:Introduction}
%%%%%%%%%%%%%%%%%%%%%%%%%%%%%%%%%

    Proprioceptive State Estimation (PropSE) refers to the process of determining the internal state or position of a robot or a robotic component (such as a limb or joint) by measuring the robot's internal properties \cite{Billard2019TrendsAnd, Diaz2023MachineLearning}. PropSE is particularly important in soft robotics, especially in terrestrial and aquatic environments, where the flexible and deformable nature of these robots makes traditional position and orientation sensing challenging \cite{Van2018SoftOptoelectronic}. During the robot's physical exchange with the external environment, the moment of touch holds the truth of the dynamic interactions \cite{Dahiya2009TactileSensing}. For most living organisms, the skin is crucial in translating material properties, object physics, and interactive dynamics via the sensory receptors into chemical signals \cite{Johansson2009CodingAnd}. When processed by the brain, they collectively formulate a \textit{feeling} of the external environment (exteroception) \cite{Sundaram2019LearningThe} and the bodily self (proprioception) \cite{Chun2021AnArtificial}. Towards tactile robotics, one stream of research aims at replicating the skin's basic functionality with comparable or superior performances \cite{Shih2020ElectronicSkins}. For example, developing novel tactile sensors \cite{Dahiya2013DirectionsToward} represents a significant research focus. Another stream of research considers robots while developing or utilizing tactile sensors \cite{OpenAI2020LearningDexterous}. It requires an interdisciplinary approach to resolve the design challenge involved \cite{Kappassov2015TactileSensing}, fostering a growing interest in tactile robotics among academia and industry \cite{Yousef2011TactileSensing}.

    We previously conducted a preliminary investigation on Vision-Based Tactile Sensing (VBTS) \cite{Wan2022VisualLearning}, which leverages the visual features of a series of soft metamaterial structures' large-scale, omni-directional adaptative deformation. The design of these metamaterial structures was subsequently generalized as a class of Soft Polyhedral Networks (SPN) \cite{Liu2024ProprioceptiveLearning}, for which high-performance proprioceptive learning in object manipulation was achieved via a node-based representation. 
    Recent literature shows the growing adoption of volumetric representation with finite element modeling as the de facto ground truth for soft, dynamic interactions \cite{Kim2020HeterogeneousSensing}. Yet, the high computational cost limits its application in robotic tasks, where real-time perception is critical \cite{Faure2012SOFA}. Aquatic machine vision remains difficult \cite{Stuart2017OceanOne} for unstructured underwater exploration with changing turbidity (relative clarity of a liquid measured by Nephelometric Turbidity Unit, or NTU). Finger-based PropSE complements aquatic machine vision by providing localized tactile perception in Simultaneous Localization and Mapping (SLAM) \cite{Suresh2021TactileSLAM}. It is a research gap to investigate the design and learning trade-off between high-fidelity proprioceptive state estimation and real-time perception in an amphibious environment \cite{Kim2020HeterogeneousSensing, Mazzeo2022MarineRobotics, Meerbeek2018SoftOptoelectronic}. In such scenarios, in-finger vision with soft robotic fingers may provide a promising solution to advance the field of tactile robotics.   

    This paper introduces a Vision-Based Tactile Sensing (VBTS) approach for real-time and high-fidelity Proprioceptive State Estimation (PropSE) with demonstrated amphibious applications in the lab and field. This is achieved using the Soft Polyhedral Network structure with marker-based in-finger vision as the soft robotic fingers for large-scale, omni-directional adaptations with amphibious tactile sensing capability. We proposed a model-based approach for PropSE by introducing rigidity-aware Aggregated Multi-Handle (AMH) constraints to optimize a volumetric parameterization of the soft robotic finger's morphological deformation. This enabled us to restructure the VBTS problem as an implicit surface model using Gaussian Processes for object shape reconstruction. We benchmarked our proposed method in shape reconstruction against existing solutions with verified superior performances. We also conducted experiments using commercial-grade motion-capture systems and touch-haptic devices, demonstrating our solution's large-scale reconstruction and touch-point estimation performances. Finally, we demonstrated the application of our proposed solutions for amphibious tactile sensing in three experiments, including a shape reconstruction experiment, a turbidity benchmarking experiment, and a tactile grasping experiment on an underwater Remotely Operated Vehicle (ROV). The following are the contributions of this study:
    \begin{itemize}
        \item Modelled Proprioceptive State Estimation (PropSE) via rigidity-aware Aggregated Multi-Handle constraints.
        \item Formulated Vision-Based Tactile Sensing (VBTS) via an Implicit Surface model for object shape reconstruction.
        \item Achieved PropSE for VBTS using Soft Polyhedral Networks with in-finger vision as robotic tactile fingertips.
        \item Benchmarked PropSE for amphibious tactile reconstruction with demonstrated applications \& testing.
    \end{itemize}    
    
    This paper is organized as follows. Section \ref{sec:LitRev} briefly reviews related literature about the role of proprioceptive state estimation in tactile robotics and its application in amphibious tactile sensing. Section \ref{sec:Method} introduces the soft robotic fingertips for this study and presents our proposed model for proprioceptive state estimation via rigidity-aware Aggregated Multi-Handle constraints. This section also formulates our proposed vision-based tactile sensing method via implicit surface modeling. All experimental results are presented in Section \ref{sec:Result}, including those for benchmarking our proposed method's performance and those conducted explicitly for amphibious tactile sensing underwater. Conclusion, limitations, and future work are enclosed in the final section, which ends this paper.

%%%%%%%%%%%%%%%%%%%%%%%%%%%%%%%%%
\section{Literature Review}
\label{sec:LitRev}
%%%%%%%%%%%%%%%%%%%%%%%%%%%%%%%%%

\subsection{Towards Dense Sensing for Tactile Robotics} 

    Tactile sensory generally involves many properties that can be digitized for robotics \cite{Li2020AReview}. For mechanics-based dynamics and control, the interactive forces and torques on the contact surface are a primary concern in robotics \cite{DeMaria2012ForceTactile}. It usually involves a certain level of material softness or structural deformation for an enhanced representation of the mechanic interactions as tactile data. The following are the three general research streams in this field. 

\subsubsection{Point-wise Sensing in 6D FT}

    Estimating forces at contact points is paramount in robotic systems, enabling awareness of physical interaction between the robot and its surrounding objects \cite{Magrini2014EstimationOf}. Robotic research, especially when dynamics and mechanics are involved, is generally more interested in utilizing the force-and-torque (FT) properties for manipulation problems by robotic hands \cite{Holladay2021PlanningFor} or locomotion tasks by legged systems \cite{Lin2020ContactSurface}. The FT properties could be succinctly represented by a 6D vector of forces and torques for a single reference point, making it comparable to the joint torque sensing in articulated robotic structures. However, the shortcut between physical contact and a point-wise 6D FT measurement may not capture the full extent of contact information for further algorithmic processing \cite{Haddadin2018TactileRobots}.
    
\subsubsection{Bio-inspired Sparse Sensing Array}
  
    Similar to the biological skin's super-resolutive mechanoreception for tactile sensing \cite{Yan2022TactileSuperResolution}, a common approach in engineering is to place an array of sensing units on the interactive surface \cite{Wu2018ASkinInspired}. Instead of going for a localized 6D force and torque contact information, researchers usually tackle the problem with enhanced pressure sensing across its entire surface from spatially distributed sensing elements \cite{Liu2022PrintedSynaptic}. As a result, one can build models or implement learning algorithms to achieve super-resolution by sampling the discrete sensory inputs. This approach continuously estimates the tactile interaction on the surface at a much higher resolution than the sensing array arrangement. Recent research \cite{Yan2021SoftMagnetic} shows that one can leverage magnetic properties to achieve de-coupled normal and shear forces with simultaneous super-resolution in tactile sensing of the normal and frictional forces for high-performing grasping.

\subsubsection{Visuo-Tactile Dense Image Sensing}

    Vision-based tactile sensing recently emerged as a popular approach to significantly increase the sensing resolution \cite{Yuan2017Gelsight}. This approach leverages the modern imaging process to visually track the deformation of a soft medium as the interface of physical interaction \cite{WardCherrier2018TacTip, Alspach2019SoftBubble}, eliminating the need for biologically inspired super-resolution \cite{Sun2022GuidingThe}. Robotic vision has already become a primary sensing modality for advanced robots \cite{Trueeb2020TowardsVisionBased}. The maturity of modern imaging technologies drives the hardware to be more compact while the software is more accessible to various algorithm libraries for real-time processing. While the high resolution of modern cameras offers significant advantages, the infinite number of potential configurations of the soft medium introduces a considerable challenge \cite{Sferrazza2019GroundTruth}.
    
\subsection{Proprioceptive State Estimation}

    For tactile applications in robotics, proprioceptive perception of joint position and body movement plays a critical role in achieving state estimation. The tactile interface is a physical separation between the intrinsic proprioception concerning the robot and the extrinsic perception concerning the object-centric environment. We focus on vision-based proprioception, which also applies to analyzing the abovementioned methods.

\subsubsection{Intrinsic Proprioception in Tactile Robotics}

    For vision-based intrinsic proprioception, the analysis is usually centered on estimating the state of the soft medium during contact, inferring tactile interaction \cite{Yamaguchi2019RecentProgress}. To establish a physical correspondence between a finite parameterization state estimation model and an infinite configuration of soft deformation \cite{Armanini2023SoftRobots}, markers that are easy to track are often used to discretize the displacement field of soft mediums. In \cite{Yamaguchi2017ImplementingTactile}, a simple blob detection method is introduced to track uniform distributed markers in a planar transparent soft layer for deformation approximation. Advanced image analysis \cite{Kroeger2016FastOptical} is also adopted to utilize machine learning algorithms to extract high-level deformation patterns from randomly spread markers over the entire three-dimensional volume of soft medium for robust state estimation \cite{Sferrazza2019DesignMotivation}. Recent research \cite{Zhang2018VisionBasedSensing} shows a promising approach to integrate physics-based models that capture the dynamic behavior of the soft medium under deformation. 
    
\subsubsection{Extrinsic Perception for Tactile Robotics}
    
    For extrinsic perception, the focus is shifted to estimating the object-level information. Tactile sensing data such as object localization, shape, and dynamics parameters could be used for task-based manipulation and locomotion \cite{Li2020AReview}. Using contact to estimate an object's global geometry is instrumental for intelligent agents to make better decisions during object manipulation \cite{Kaboli2019TactileBasedActive}. Usually, tactile sensing is employed for estimating the object's shape in visually occluded regions, thus playing a complementary role to vision sensors \cite{Ilonen2014ThreeDimensionalObject, Wang20183DShape}. However, in scenarios where a structured environment with reliable external cameras is unavailable or impractical, such as during exploration tasks in unstructured environments, tactile sensing can provide valuable feedback to achieve environmental awareness \cite{Yasa2023AnOverview}.
        
\subsection{Amphibious Tactile Robotics}

    Amphibious environments present a unique and dynamic challenge for robotic systems \cite{Baines2022MultiEnvironmentRobotic}. Robots operating in these environments must contend with vastly different physical properties, including changes in buoyancy, friction, and fluid dynamics \cite{Rafeeq2021LocomotionStrategies}. Furthermore, the transition between water and air requires robots to adapt their sensory systems and control strategies to function effectively in each medium \cite{Yu2012OnA}. 

    Developing effective tactile sensors for amphibious robots presents several challenges. Sensors must be robust enough to withstand the harsh aquatic environment and be sensitive enough to detect subtle changes in water and air \cite{Subad2021SoftRobotic}. The transition between these two media can also cause sensor drift and require calibration to maintain accuracy \cite{Li2023AnAerial}. Despite these challenges, there are exciting opportunities in amphibious tactile robotics, with improved sensitivity, durability, and resistance to environmental factors \cite{Aggarwal2015HapticObject}. However, a research gap remains in developing an effective tactile sensing method with an integrated finger-based design that directly applies to amphibious applications.

%%%%%%%%%%%%%%%%%%%%%%%%%%%%%%%%%
\section{Materials \& Methods}
\label{sec:Method}
%%%%%%%%%%%%%%%%%%%%%%%%%%%%%%%%%
\subsection{Soft Polyhedral Network with In-Finger Vision}
\label{sec:Method-Design}
%%%%%%%%%%%%%%%%%%%%%%%%%%%%%%%%%

\begin{figure}[!h]
        \centering
        \includegraphics[width=0.8\textwidth]{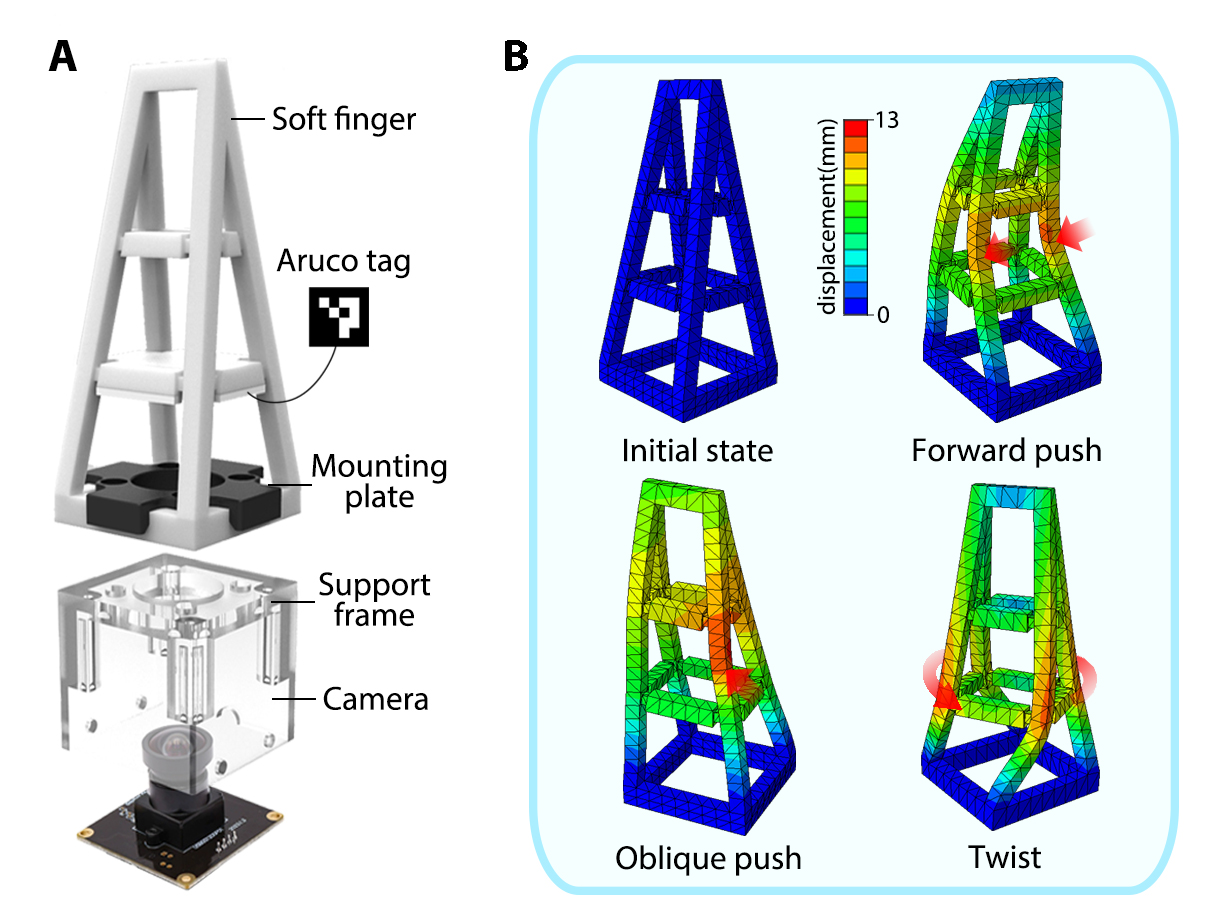}
        \caption{
        \textbf{Assembly and omni-adaptive capability of the soft finger.} 
        (A) The assembly consists of a soft finger, a rigid plate pasted with an ArUco tag, a mounting plate, a support frame, and a camera. 
        (B) The finger deformation by forward push, oblique push, and twist shows the omni-adaptive capability.
        }
        \label{fig:Method-Design}
    \end{figure}

    Soft grippers can achieve diverse and robust grasping behaviors with a relatively simple control strategy \cite{Yang2023DynamicCapture}. In this study, we adopted our previous work in a class of Soft Polyhedral Networks with in-finger vision as the soft robotic finger \cite{Wan2022VisualLearning, Liu2024ProprioceptiveLearning}. As shown in Fig. \ref{fig:Method-Design}A, the specific design is modified using an enhanced mounting plate to fix the soft finger and made waterproof for amphibious tactile sensing. The soft finger features a shrinking cross-sectional network design towards the tip, capable of omni-directional adaptation during physical interactions, as shown in Fig. \ref{fig:Method-Design}B. We fabricated the finger by vacuum molding using Hei-cast 8400, a three-component polyurethane elastomer. Based on our previous work, we mixed the three components with a ratio of 1:1:0, producing a hardness of 90 (Type A) to achieve reliable spatial adaptation for grasping. 

    An ArUco\footnote{\url{http://sourceforge.net/projects/aruco/}} tag \cite{Garrido2014AutomaticGeneration} is attached to the bottom side of a rigid plate mechanically fixed with the four lower crossbeams of the soft finger. A monocular RGB camera with a field of view (FOV) of 130$^{\circ}$ is fixed at the bottom inside a transparent support frame as in-finger vision, video-recording in a high frame rate of 120 FPS (frames per second) at 640 × 480 pixels resolution. When the soft robotic finger interacts with the external environment, live video streams captured by the in-finger vision camera provide real-time pose data of the ArUco tag as rigid-soft kinematics coupling constraints for the PropSE of the soft robotic finger. This marker-based in-finger vision design is equivalent to a miniature motion capture system, efficiently converting the soft robotic finger's spatial deformation into real-time 6D pose data.
    
%%%%%%%%%%%%%%%%%%%%%%%%%%%%%%%%%
\subsection{Volumetric Modeling of Soft Deformation for PropSE}
\label{sec:Method-ModelPropSE}
%%%%%%%%%%%%%%%%%%%%%%%%%%%%%%%%%

    Our proposed solution begins by formulating a volumetric model of the soft robotic finger in a 3D space $\mathbf{\Omega}\in{\mathbb{R}^{3}}$ filled with homogeneous elastic material. The distribution of the internal elastic energy within the volumetric elements varies significantly depending on the boundary conditions defined. The PropSE process requires an accurate determination of a smooth deformation map, $\Phi:\mathbf{\Omega}\rightarrow{\tilde{\mathbf{\Omega}}}$, that facilitates the geometric transformation of the soft body from its initial state, represented by $\mathbf{\Omega}$, to a deformed state, denoted as $\tilde{\mathbf{\Omega}}$. This transformation is characterized by minimizing a form of variational energy measuring the distortion of the soft body \cite{Lanczos1986TheVariational}. As a result, the PropSE performance depends on finite element discretization and the choice of energy function that characterizes deformation.

\subsubsection{Volumetric Parameterization of Whole-body Deformation}
\label{sec:Method-ModelPropSE-Discretize}

\begin{figure}[htbp]
        \centering
        \includegraphics[width=0.8\textwidth]{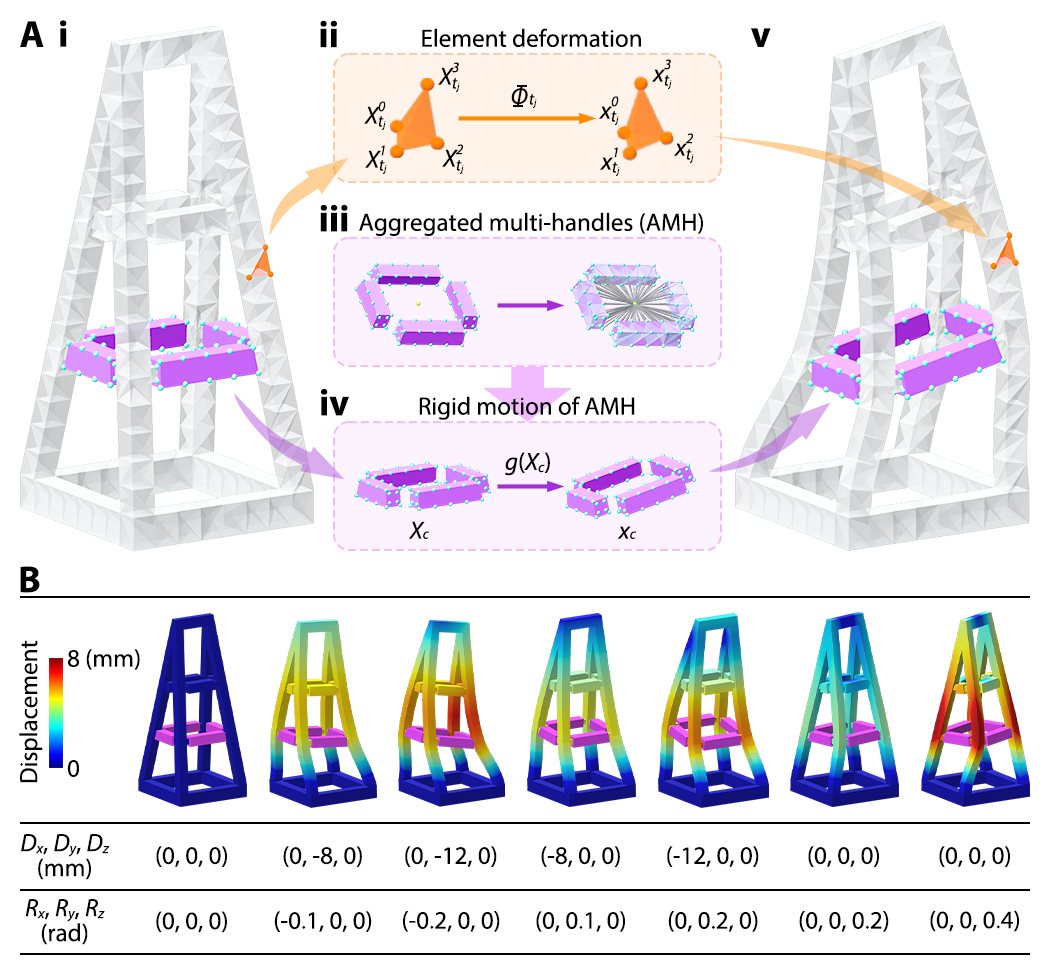}
        \caption{
        \textbf{Proprioceptive deformation modeling and estimation of Omni-Adaptive Soft Finger.} 
        (A) Representation of the proprioceptive model, including
            i) Initial undeformed configuration $\Omega$ of the soft finger, discretized using tetrahedral mesh;
            ii) Local affine mapping $\Phi_{t_j}$ applies on $t_j$ element, transforming each vertex from $\mathbf{X}_{t_{j}}^i\in{\mathbb{R}^3}$ to $\mathbf{x}_{t_{j}}^i\in{\mathbb{R}^3}$,$i\in{\{1,2,3,4\}}$;
            iii) Approximation of visual observed marker area as Aggregated Multi-Handles (AMH) on the tetrahedral mesh (xx colored); 
            iv) Applies uniform rigid motion $g\in{SE(3)}$ on all AMH that drives soft finger to a deformed configuration $\tilde{\mathbf{\Omega}}$. 
        (B) Demonstration of soft finger deformation reconstructions under a series of rigid motions applied on AMH, including bending and twisting.
        }
        \label{fig:Method-ModelPropSE-Discretize}
    \end{figure}

    We denote a tetrahedral mesh of the discretized soft body using $\mathcal{M}=\{\mathcal{V}, \mathcal{T}\}$, where $\mathcal{V}=\{\mathbf{x}_1, ..., \mathbf{x}_n\}$ is the set of vertices $\mathbf{x}_{i}\in{\mathbb{R}^3}$, and $\mathcal{T}=\{t_1, ..., t_m\}$ is the set of tetrahedra elements, as shown in Fig. \ref{fig:Method-ModelPropSE-Discretize}A(i). 

    When the soft body deforms, a collection of chosen linear approximated local deformation maps are applied to $\mathcal{M}$ over each tetrahedron element $t_{j}$ via an affine transformation:
    \begin{equation}\label{piecewiseAFF}
        \Phi|_{t_j}(\mathbf{X})=\mathbf{A}_{t_{j}}\mathbf{X}+\mathbf{b}_{t_{j}}
        ,
    \end{equation}
    where $\mathbf{X}\in{\mathbb{R}^3}$ stands for all points inside element $t_j$, $\mathbf{A}_{t_{j}}\in{\mathbb{R}^{3\times{3}}}$ is the differential part of the deformation map, and $\mathbf{b}_{t_{j}}\in{\mathbb{R}^{3}}$ is the translational part. We choose this piecewise linear deformation map for computational efficiency. High-order deformation functions can be used for better approximation if needed \cite{Longva2020HigherOrderFinite}.

    As shown in Fig. \ref{fig:Method-ModelPropSE-Discretize}A(ii), for any $t_j$ element, the local affine transformation applied on each vertex is denoted as:
    \begin{equation}\label{affine_form}
        \begin{bmatrix}
            \!\mathbf{A}_{t_{j}} & \!\!\!\!\mathbf{b}_{t_{j}}\!
            \end{bmatrix}
            \!\!\cdot\!\!{
            \begin{bmatrix}
            \!\mathbf{X}_{t_{j}}^1 
            &\!\!\!\!\mathbf{X}_{t_{j}}^2
            &\!\!\!\!\mathbf{X}_{t_{j}}^3
            &\!\!\!\!\mathbf{X}_{t_{j}}^4\!\\
            \!\mathbf{1}
            &\!\!\!\!\mathbf{1}
            &\!\!\!\!\mathbf{1}
            &\!\!\!\!\mathbf{1}\!\end{bmatrix}}
        =
        \begin{bmatrix}
        \mathbf{x}_{t_{j}}^1 
        &\!\!\!\!\mathbf{x}_{t_{j}}^2
        &\!\!\!\!\mathbf{x}_{t_{j}}^3
        &\!\!\!\!\mathbf{x}_{t_{j}}^4\!
        \end{bmatrix}
        ,
    \end{equation}
    where $\mathbf{x}_{t_{j}}^i\in{\mathbb{R}^3}$, $i\in{\{1,2,3,4\}}$ are the deformed vertices location of $t_j$ tetrahedron, and $\mathbf{X}_{t_{j}}^i\in{\mathbb{R}^3}$ are the corresponding initial vertices location. 

    Therefore, the deformation gradient $\mathbf{A}_{t_{j}}$ in the chosen piecewise linear transformation in Eq. \eqref{piecewiseAFF} can be expressed as a linear combination of unknown deformed element vertices location $\mathbf{x}_{t_{j}}$ using the following formulation:
    \begin{equation}
        \mathbf{A}_{t_{j}}(\mathbf{x}_{t_{j}})
        =
        \frac{\partial{\Phi|_{t_j}}}{\partial{\mathbf{X}}}
        =
        \mathbf{D}_s(\mathbf{x}_{t_{j}})\cdot{\mathbf{D}_m^{-1}(\mathbf{X}_{t_{j}})}
        ,
    \end{equation}
    where
    \begin{equation}
        \mathbf{D}_s(\mathbf{x}_{t_{j}}) 
        = \begin{bmatrix}
        \mathbf{x}_{t_{j}}^2-\mathbf{x}_{t_{j}}^1 
        &\mathbf{x}_{t_{j}}^3-\mathbf{x}_{t_{j}}^1
        &\mathbf{x}_{t_{j}}^4-\mathbf{x}_{t_{j}}^1
        \end{bmatrix}
        ,
    \end{equation}
    \begin{equation}
       \mathbf{D}_m(\mathbf{X}_{t_{j}}) 
       = \begin{bmatrix}
       \mathbf{X}_{t_{j}}^2-\mathbf{X}_{t_{j}}^1 
       &\mathbf{X}_{t_{j}}^3-\mathbf{X}_{t_{j}}^1
       &\mathbf{X}_{t_{j}}^4-\mathbf{X}_{t_{j}}^1
       \end{bmatrix}
       .
    \end{equation}
    
    For a discretized tetrahedral mesh $\mathcal{M}$, the collection of deformation maps $\{\Phi_{t_{j}}\}_{t_{j}\in{\mathcal{T}}}$ for all tetrahedra elements should uniquely determine the deformed shape of the soft body \cite{Aigerman2013Injective}. 

\subsubsection{Geometry-Related Deformation Energy Function}
\label{sec:Method-ModelPropSE-EnergyFunc}

    To mimic the physical deformation behavior, the specific energy function form of the deformation map $\Psi(\Phi_{t_{j}})$ needs to be specified. Several formulations of geometry-related deformation energies, such as As-Rigid-As-Possible (ARAP) \cite{Liu2008ALocalGlobal}, conformal distortion \cite{Aigerman2015SeamlessSurface}, and isometric distortion \cite{Smith2015BijectiveParameterization}, have been proposed in recent literature. 
    
    Instead of deriving the energy of the system explicitly using constitutive relation and balance equations \cite{Lai2009IntroductionTo}, we choose a symmetric Dirichlet form of energy function \cite{Rabinovich2017ScalableLocally} to characterize the deformation, which indicates isometric distortion and behaves well in the case of our soft finger. Since the deformation should be irrelevant to the translation, the discrete element energy function only takes the gradient augment of each deformation maps $\{\Phi_{t_{j}}\}_{t_{j}\in{\mathcal{T}}}$ as:
    \begin{equation}\label{eq:sd_element_energy}
        \Psi(\Phi_{t_{j}})=\Psi{(\mathbf{A}_{t_{j}})}=||\mathbf{A}_{t_{j}}||^2_\mathcal{F}+||\mathbf{A}_{t_{j}}^{-1}||^2_\mathcal{F}
        ,
    \end{equation}
    where $||\cdot||_\mathcal{F}$ is the Frobenius norm. The accumulated discrete element energy functional of the soft body denotes:
    \begin{equation}\label{iso-energy}
        E(\mathbf{x})=\sum_{t_j\in{\mathcal{T}}}{\Psi{(\mathbf{A}_{t_{j}}(\mathbf{x}))}}
        ,
    \end{equation}
    where $\mathbf{x}\in{\mathbb{R}^{3\times{n}}}$ contains all discretized vertices location of the soft body $\mathcal{M}$.

\subsubsection{Rigidity-Aware Aggregated Multi-Handles Constraints}
\label{sec:Method-ModelPropSE-RigidityAwareAMH}

    Monocular cameras are generally considered the primary sensory for environmental perception due to their ease of use and availability, compared to multi-view systems. However, deformable shape reconstruction from 2D image observations is well-known as an ill-posed inverse problem and has been actively researched \cite{Tretschk2023StateOf}. We leverage the proposed volumetric discretized model and introduce rigidity-aware Aggregated Multi-Handle (AMH) constraints to make this problem trackable, aiming at reconstructing the soft finger's deformed shape reliably.
    
    We model the mechanical coupling of the rigid plate for the fiducial marker in Fig. \ref{fig:Method-Design}A as a uniform rigid transformation $g$ for each attached node in the discrete model $M$, as shown in Figs. \ref{fig:Method-ModelPropSE-Discretize}A(iii)\&(iv):
    \begin{equation}\label{amh-constrains}
        \mathbf{x}_{h}=g(\mathbf{X}_h)
        ,
    \end{equation}
    where $\mathbf{x}_h\in{\mathbb{R}^{3\times{p}}}$ contains deformed location of $p$ vertices related to the rigidity-aware AMH constraints while $\mathbf{X}_h\in{\mathbb{R}^{3\times{p}}}$ contains the corresponding undeformed vertices location. The rigid transformation $g$ is estimated by fiducial markers widely used in robotic vision.

\subsubsection{Geometric Optimization for Shape Estimation}
\label{sec:Method-ModelPropSE-ShapeEst}

    With the discrete energy function Eq. \eqref{iso-energy} of the given soft body $\mathcal{M}$ and observed kinematics constraints Eq. \eqref{amh-constrains}, soft body shape estimation can be directly translated into a constrained geometry optimization problem as:
    \begin{equation}\label{eq:prime_fun}
        \begin{aligned}
        \min_{\mathbf{x}} \quad & \sum_{t_j\in{\mathcal{T}}}{\Psi{(\mathbf{A}_{t_{j}}(\mathbf{x}))}}
        ,\\
        \textrm{s.t.} \quad & \mathbf{x}_{h}=g(\mathbf{X}_h)
        .\\
        \end{aligned}
    \end{equation}
    
    Instead of considering kinematics constraints as hard boundary conditions, we enforce them by appending quadratic penalty terms to $E(\mathbf{x})$ in Eq. \eqref{iso-energy} for easier handling, which results in
    \begin{equation}\label{constr_fun}
        \Tilde{E}(\mathbf{x}) = \sum_{t_j\in{\mathcal{T}}}{\Psi{(\mathbf{A}_{t_{j}}(\mathbf{x}))}+\omega||\mathbf{x}_h-g(\mathbf{X}_h)||^2}
        .
    \end{equation}
    As illustrated in Fig. \ref{fig:Method-ModelPropSE-Discretize}A(v), we can achieve deformed shape estimation by minimizing the augmented energy function in Eq. \eqref{constr_fun} as:
    \begin{equation}\label{min-func}
        \mathbf{x}^{*}=
        \mathop{\arg\min}\limits_{\mathbf{x}}\Tilde{E}(\mathbf{x};\omega,g)
        ,
    \end{equation}
    where $\omega$ is the penalty parameter for the corresponding unconstrained minimization problem. A greater penalty weight will lead to better constraint satisfaction but poorer numerical conditions.
    
    In practice, we set $\omega = 10^5$ and compute the deformed vertices positions $\mathcal{V}$ by iteratively minimizing Eq. \eqref{min-func} using a Newton-type solver shown in Alg. \ref{pn}. As shown in Fig. \ref{fig:Method-ModelPropSE-Discretize}B, a series of physically plausible deformations of the soft finger under observed constraints are reconstructed in real time using our proposed optimization approach. 

    \begin{algorithm}[!t]
      \caption{Projected Hessian Algorithm }\label{pn}
      \begin{algorithmic}[1]
        \State \textbf{Input}: Rigid transformation of AMH $g$
        \State \textbf{Output}: Estimated positions of deformed vertices $\mathbf{x}^{*}$
        \Require
          \Statex Vertices positions of current shape $\mathbf{x}_0$
          \Statex Convergence tolerance $\epsilon$
          \Statex Maximum number of iterations $N_{\text{max}}$
            \State $k \gets 0$
            \State Compute gradient $d_{k} = \nabla \Tilde{E} (\mathbf{x}_k)$
            \State And Hessian $H_k = \nabla^2 \Tilde{E}(\mathbf{x}_k)$
        \While{$\|d_{k}\| > \epsilon$ and $k < N_{\text{max}}$}
          \State Solve $H_k \Delta \mathbf{x}_k = - d_k$ for $\Delta \mathbf{x}_k$
          \State Project $\Delta \mathbf{x}_k$ onto the feasible region
          \State Update iterate: $\mathbf{x}_{k+1} \gets \mathbf{x}_k + \Delta \mathbf{x}_k$
          \State $k \gets k + 1$
        \EndWhile
            \State iteration stop $\mathbf{x}^{*} = \mathbf{x}_k$
      \end{algorithmic}
    \end{algorithm}

%%%%%%%%%%%%%%%%%%%%%%%%%%%%%%%%%
\subsection{Object Shape Estimation using Tactile Sensing}
\label{sec:Method-ModelGPIS}
%%%%%%%%%%%%%%%%%%%%%%%%%%%%%%%%%

    While proprioception refers to being aware of one's movement, tactile sensing involves gathering information about the external environment through the sense of touch. This section presents an object shape estimation approach by extending the PropSE method proposed in the previous section to tactile sensing.

    Since our soft finger can provide large-scale, adaptive deformation conforming to the object's geometric features through contact, we could infer shape-related contact information from the finger's estimated shape during the process. We assume the soft finger's contact patch coincides with that of the object during grasping. As a result, we can predict object surface topography using spatially distributed contact points on the touching interface.

    \begin{figure*}[t]
        \centering
        \includegraphics[width=\textwidth]{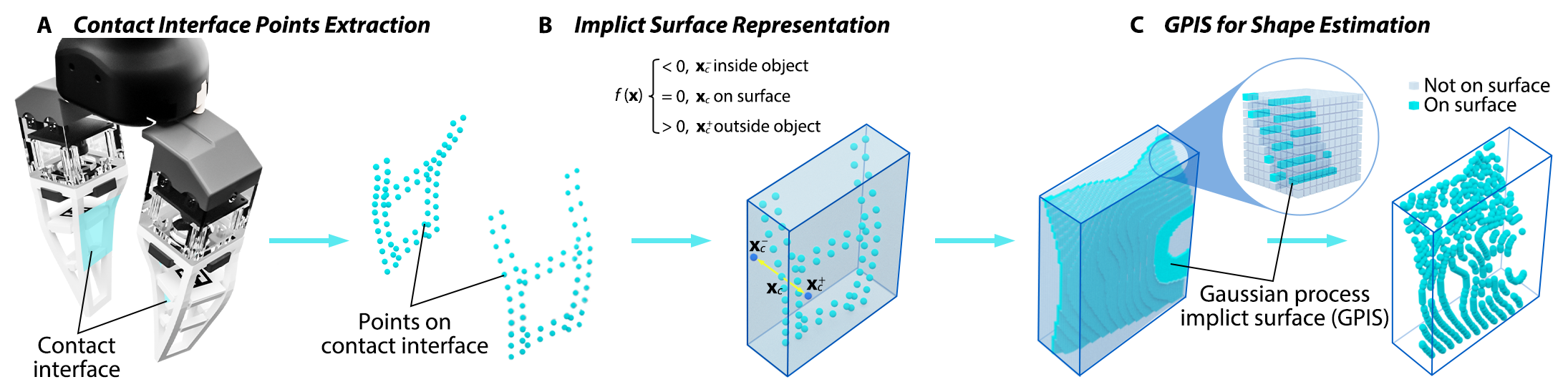}
        \caption{
        \textbf{Pipeline for contact interface geometry sensing using deformed positions of soft finger mesh nodes.} 
        (A) Because the soft finger can deform and adapt its shape to fit the contours of the object being grasped, we take the deformed soft finger mesh nodes as approximate multi-contact points on the contact interface. 
        (B) In addition to the mesh nodes $\mathbf{x}_{c}$ on the contact interface, auxiliary training points $\mathbf{x}^{-}_{c}$ and $\mathbf{x}^{+}_{c}$ are generated in this step to increase the accuracy of the implicit surface reconstruction. 
        (C) Gaussian process implicit surface model is adopted for contact object surface patch estimation.
        }
        \label{fig:Method-ModelGPIS}
    \end{figure*}
    
\subsubsection{Contact Interface Points Extraction}

    Based on the spatial discretization model in Section \ref{sec:Method-ModelPropSE-Discretize}, an indexed set $\mathcal{I}=\{c_1,c_2,...,c_k\}$ of nodes located at the upper area of the soft finger mesh $\mathcal{M}$ are extracted as contact interface points, as shown in Fig. \ref{fig:Method-ModelGPIS}A.

    With each of the observed AMH constraints input, we could determine the positions of these contact interface points by first solving Eq. \eqref{min-func}, then extracting corresponding nodes using indexed set $\mathcal{I}$ by solving the deformed positions of vertices $\mathcal{V}$: $\mathbf{x}_c=\{\mathbf{x}_{i}| \mathbf{x}_{i}\in{\mathcal{V}}, i\in{\mathcal{I}} \}$.
    
\subsubsection{Implict Surface Representation for Object Shape} 

    Considering the grasping action using a soft finger as a multi-point tactile probe, the object surface patches could be progressively reconstructed by these gripping actions with collected positions of contact interface points $\mathbf{x}_{c}$ extracted from the soft finger.
    
    An implicit surface representation is defined by a function that can be evaluated at any point in space, yielding a value indicating whether the point is inside the object, outside the object, or on the object's surface. For the 3-D space considered in our problem, this function $f:\mathbb{R}^{3}\rightarrow{\mathbb{R}}$ is defined as:
    \begin{equation}\label{imp_surf}
            f(\mathbf{x})
            \begin{cases}
                <0, & \mbox{ if $\mathbf{x}$ inside the object};\\
                =0, & \mbox{ if $\mathbf{x}$ on the surface};\\
                >0, & \mbox{ if $\mathbf{x}$ outside the object}.
            \end{cases}
    \end{equation}

    As is shown in Fig. \ref{fig:Method-ModelGPIS}B, we only collected positions of partial contact interface points $\mathbf{x}_c$, which are assumed to coincide with the object surface for each gripping action. While surface points are observed, we do not explicitly observe off-surface or internal points exemplars. For those unobserved cases in Eq. \eqref{imp_surf}, we generate control points of the corresponding two types to express the directional information of the surface using the method described in \cite{GerardoCastro2015LaserRadarData}.

\subsubsection{GPIS for Surface Estimation}
    
    An object's shape is estimated by finding the points with zero value of implicit surface function Eq. \eqref{imp_surf} (i.e., the isosurface) in the 3D region of interest. The Gaussian Process Implicit Surface (GPIS) method can be used as a tool for object surface reconstruction from partial or noisy 3D data. It is a non-parametric probabilistic method often used for tactile and haptic exploration \cite{Dragiev2011GaussianProcess, Ottenhaus2016LocalImplicit}. 
    
    A Gaussian Process (GP) is a collection of $N$ random variables with a joint Gaussian distribution which can be specified using its mean and covariance functions. The collected contact interface point and the generated control point positions $\mathcal{X}=\{\mathbf{x}_{1},\mathbf{x}_{2},...,\mathbf{x}_{N}\}$ for each grasping action and the corresponding observed values are denoted as $\mathcal{Y}=\{\mathbf{y}_{1},\mathbf{y}_{2},...,\mathbf{y}_{N}\}$. Here, $\mathbf{y}_{i} = f(\mathbf{x}_{i})+\epsilon$, where $\epsilon\sim{\mathcal{N}(0,\sigma^{2}_{\epsilon})}$ denotes Gaussian noise with zero mean and  $\sigma^{2}_{\epsilon}$ variance. As a result, the GP can be written as $f(\mathbf{x})\sim{\mathcal{GP}(m(\mathbf{x}),k(\mathbf{x},\mathbf{x}^{\prime}))}$, where $m(\mathbf{x})$ is the mean function and $k(\mathbf{x},\mathbf{x}^{\prime})$ is the covariance function \cite{Rasmussen2005GaussianProcesses}. 
    
    In our implementation, we used the radial basis function kernel, which is characterized by the two hyper-parameters, the variance $\sigma^{2}_{f}$ and the length scale $l$, expressed as the following:
    \begin{equation}\label{SE_COV}
        k(\mathbf{x},\mathbf{x}^{\prime})=\sigma^{2}_{f}\exp{(-\frac{||\mathbf{x}-\mathbf{x}^{\prime}||^{2}}{2l^{2}})}
        .
    \end{equation}

    With the covariance function and the observation data, the predictive mean $\Bar{f}(\mathbf{x}^{*})$ and variance $\Bar{\mathcal{V}}(\mathbf{x}^{*})$ at a query point $\mathbf{x}^{*}$ are:
    \begin{equation}\label{pred_mean}
        \Bar{f}(\mathbf{x}^{*}) = \mathbb{E}[f(\mathbf{x}^{*})|\mathcal{X},\mathcal{Y},\mathbf{x}^{*}]=k(\mathcal{X},\mathbf{x}^{*})^{\rm T}\Sigma \mathcal{Y}\\
        ,
    \end{equation}
    \begin{equation}\label{variance}
        \Bar{\mathcal{V}}(\mathbf{x}^{*}) = k(\mathbf{x}^{*},\mathbf{x}^{*})-k(\mathbf{x}^{*},\mathbf{x})^{\rm T}\Sigma k(\mathbf{x}^{*},\mathbf{x})
        ,
    \end{equation}
    where $\Sigma=(k(\mathcal{X},\mathcal{X})+\sigma^{2}_{\epsilon}\mathcal{I})^{-1}$. After voxelizing the bounding box volume enclosing the partially deformed finger-object interface, the zero-mean isosurface can be extracted from posterior estimation, which approximates the local shape of a grasped object, as is shown in Fig. \ref{fig:Method-ModelGPIS}C.

%%%%%%%%%%%%%%%%%%%%%%%%%%%%%%%%%
\section{Results}
\label{sec:Result}
%%%%%%%%%%%%%%%%%%%%%%%%%%%%%%%%%
\subsection{On Vision-based Proprioceptive State Estimation}
\label{sec:Result-VBPropSE}
%%%%%%%%%%%%%%%%%%%%%%%%%%%%%%%%%

    Here, we first present the benchmarking results against two widely adopted methods to demonstrate the superior performance of our proposed vision-based PropSE method. Then, we present the results of our proposed vision-based PropSE method using two experiment setups. One leverages motion capture markers as ground truth, providing high-precision but sparse measurements. The other uses a touch-haptic device for ground truth data collection, which is less accurate but contains larger measuring coverage on the soft finger.    
        
    The implementation of the proposed geometric optimization-based algorithm (Alg. \ref{pn}) was developed in C++ and evaluated on a computer with an Intel Core\textsuperscript{\texttrademark} i7 3.8 GHz CPU and 16 GB of RAM. By leveraging the capabilities of algorithmic differentiation within the numerical solver, \textit{Eigen} \cite{Peltzer2020EigenAD}, this system demonstrated the capability to compute deformations involving 1,500 tetrahedra in real-time, achieving frame rates up to 20 fps.
    
\subsubsection{Comparison with the Conventional Methods}
\label{sec:Result-VBPropSE-Compare}

    We performed a comparative analysis with two widely adopted techniques to showcase the efficacy of our shape estimation method. One is Abaqus, a premier finite element analysis (FEA) software extensively applied in structural analysis and deformation modeling across various engineering disciplines. This comparison aims to highlight the versatility and precision of our approach within contexts requiring intricate modeling capabilities. (Please refer to Appendix A for further details concerning the Abaqus simulation.)

    The other is the As-Rigid-As-Possible (ARAP) method \cite{Fang2020KinematicsOf}, a widely adopted method in digital geometry processing for estimating object shapes through minimal rigid deformation. This comparison is particularly valuable, as ARAP's principles of shape preservation align closely with the core objectives of our shape estimation task, providing valuable benchmarking. (Please refer to Appendix B for further details regarding our implementation.)
    
    Table \ref{tab:Result-VBPropSE-Compare} compares our proposed method's run time and mean error with those mentioned earlier. Each method is evaluated on five meshes with increasing resolutions, resulting in 1k, 1.5k, 3k, 6k, and 12k elements. The soft finger underwent six motions applied on the AMH shown in Fig. \ref{fig:Method-ModelPropSE-Discretize}B with all the deformation data recorded. We treat the results from Abaqus as the ground truth. Results show that our method is 40 to 700 times faster than Abaqus and 1 to 2 times faster than ARAP at different resolutions. We also compared the mean errors of all nodes estimated by our method and ARAP when benchmarked against Abaqus. The results show that our method's mean error decreases significantly, from 0.346 mm to 0.086 mm, as the number of elements increases. The ARAP's error ranges from 0.7 mm to 1.0 mm for different meshes. Our approach shows significant advantages over Abaqus and ARAP regarding running time and accuracy.

    \begin{table}[t!]
        \caption{\textsc{Run Time and Mean Error Comparisons of Abaqus, ARAP, and Our Method.}}
        \centering
        \setlength{\tabcolsep}{1.3mm}{
            \begin{threeparttable}
                \begin{tabular}{ccccccc}
                \hline
                \multicolumn{2}{c}{Number of Elements}              & 1k     & 1.5k   & 3k     & 6k    & 12k   \\ \hline
                \multirow{3}{*}{Run Time (s)}           & Abaqus    & 16.7   & 17.1   & 18.7   & 22.1  & 27.8  \\
                                                        & ARAP      & 0.0485 & 0.0776 & 0.151  & 0.304 & 0.703 \\
                                                        & Ours      & 0.0231 & 0.0452 & 0.0905 & 0.241 & 0.639 \\ \hline
                \multirow{3}{*}{Mean Error$^{*}$ (mm)}  & Abaqus    & NA     & NA     & NA     & NA    & NA    \\
                                                        & ARAP      & 0.857  & 0.902  & 0.754  & 0.878 & 0.926 \\
                                                        & Ours      & 0.346  & 0.149  & 0.130  & 0.143 & 0.086 \\ \hline
                \end{tabular}
                \begin{tablenotes}
                \footnotesize
                \item $^{*}$: The benchmark of mean error is Abaqus.
                \end{tablenotes}
            \end{threeparttable}
        }
        \label{tab:Result-VBPropSE-Compare}
    \end{table}

    The optimization solver deployed to minimize the ARAP energy leverages the local/global method (as detailed in Appendix B). While this solver efficiently approximates the local minimum, its approach to convergence towards a numerical minimum necessitates a considerable number of iterations, a characteristic underscored during implementation \cite{Rabinovich2017ScalableLocally}. We fixed the number of iterations at 10 for our benchmarking procedure to achieve convergence. This predefined iteration limit could account for the observed comparative slowness of the ARAP optimization solver relative to our proposed method. Regarding the evaluation of mean error, the suboptimal performance of ARAP, as compared to ours, might be attributed to the local/global optimization solver settings. Moreover, the deformation energy model used by ARAP might not fully encompass the non-linear deformation behaviors of our soft robotic fingers.
       
    We also observe that the error of our method decreases most dramatically when the number of elements increases from 1k to 1.5k, and the error reduction from 1.5k to 6k is marginal. Hence, the mesh with 1.5k elements is the most appropriate for our method, achieving both faster run speed and minor error, which was selected for real-time estimation in the following experiments. (Please refer to Appendix C for additional results on Algorithm \ref{pn} parameter.)
    
\subsubsection{Deformation Estimation with Motion Capture Markers} 
\label{sec:Result-VBPropSE-MoCap}

\begin{figure}[!t]
        \centering
        \includegraphics[width=0.8\textwidth]{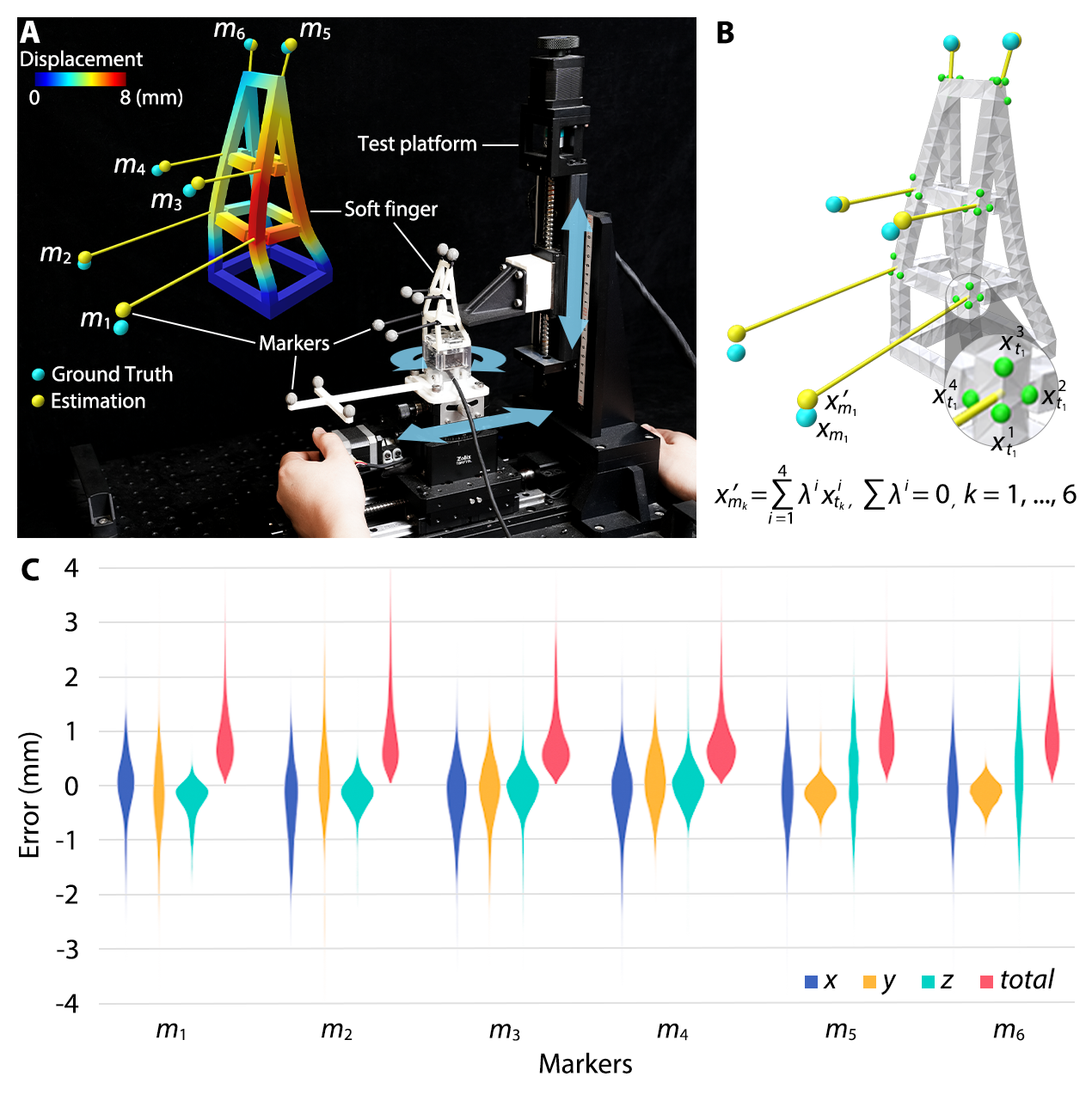}
        \caption{
        \textbf{Estimated marker deformation obtained by proposed proprioceptive state estimation method.} 
        (A) Experimental setup, including the soft finger, embedded with an RGB camera, a manual three-axis motion test platform, and six motion capture markers $m_1, m_2, ..., m_6$, rigidly attached to the soft finger. 
        (B) The estimated position of the marker $x_{m_{k}}^{\prime}$ is calculated using the barycentric coordinate of the corresponding attached tetrahedron $t_k$, while the ground truth reading $x_{m_{k}}$ is obtained from the motion capture system. 
        (C) The corresponding error for each marker's three-dimensional deformation and total norm.
        }
        \label{fig:Result-VBPropSE-MoCap}
    \end{figure}

    Shown in Fig. \ref{fig:Result-VBPropSE-MoCap}A is the soft robotic finger mounted on a three-axis motion platform for interactive deformation estimation. The test platform is operated manually to generate a set of contact configurations between the soft finger and the indenter. During the process, the in-finger camera streams real-time image data at a resolution of 640 × 480 pixels. Using an off-the-shelf ArUco detection library, the detected AMH rigid motion is fed into our implemented program for deformation estimation.
   
    A motion capture system (Mars2H by Nokov, Inc.) was used to track finger deformations through nine markers with an 8 mm radius. Among them, six markers were divided into three pairs, which were rigidly attached to the fingertip ($m_5, m_6$), the first layer ($m_3, m_4$), and the second layer ($m_1, m_2$) of the soft finger, respectively. The other three markers were attached to the platform and used as the reference reading to align the motion capture system's reference frame with the platform's coordinate frame.

    The markers were attached to the soft finger with rigid links, as shown in Fig. \ref{fig:Result-VBPropSE-MoCap}B. We designed the connecting links to be three lengths to avoid occlusion during tracking. We assume each marker is rigidly attached to the nearest tetrahedron on the parameterized mesh model $\mathcal{M}$, representing the estimated marker location using barycentric coordinates of the corresponding tetrahedron element in the soft robotic finger's deformed states:
    \begin{equation}\label{barycentric}
        \mathbf{x}_{m_{k}}^{\prime} = \sum^{4}_{i=1}\bm{\lambda}^{i}_{t_k}\cdot{\mathbf{x}_{t_{k}}^{i}},k\in{\{1,2,...,6\}},
    \end{equation}
    
    \begin{equation}\label{barycentric_lamba}
        \sum^{4}_{i=1}\bm{\lambda}^{i}_{t_k} = 1, t_{k}\in{\mathcal{T}}.
    \end{equation}
    
    Due to the rigid connection assumption, the barycentric coordinates $\bm{\lambda}_{t_{k}}$ are constant during deformation. We solve the barycentric coordinates in Eq. \eqref{barycentric} using the tetrahedron's initial vertex position and corresponding tracked marker position without contact. The marker position prediction model is a linear combination of the deformed vertex position of the corresponding tetrahedron resulting from geometric optimization in Alg. \ref{pn} using calibrated barycentric coefficients. (See Movie S1 in the Supplementary Materials for a video demonstration.)
    
    We visualize the error distribution with 3k pairs of the six markers' estimated and ground truth positions as illustrated in Fig. \ref{fig:Result-VBPropSE-MoCap}C. The norm of the six markers' total error is within 3 mm, while error distribution along each axis is centered around the $(-2, 2)$ mm range. As the marker prediction model in Eq. \eqref{barycentric} comprises calibration and geometric optimization, the error distribution of six sparse markers may only partially validate the proposed method, leading to the next experiment.
    
\subsubsection{Deformation Estimation using Touch Haptic Device} 
\label{sec:Result-VBPropSE-Haptic}

\begin{figure*}[htbp]
        \centering
        \includegraphics[width=\textwidth]{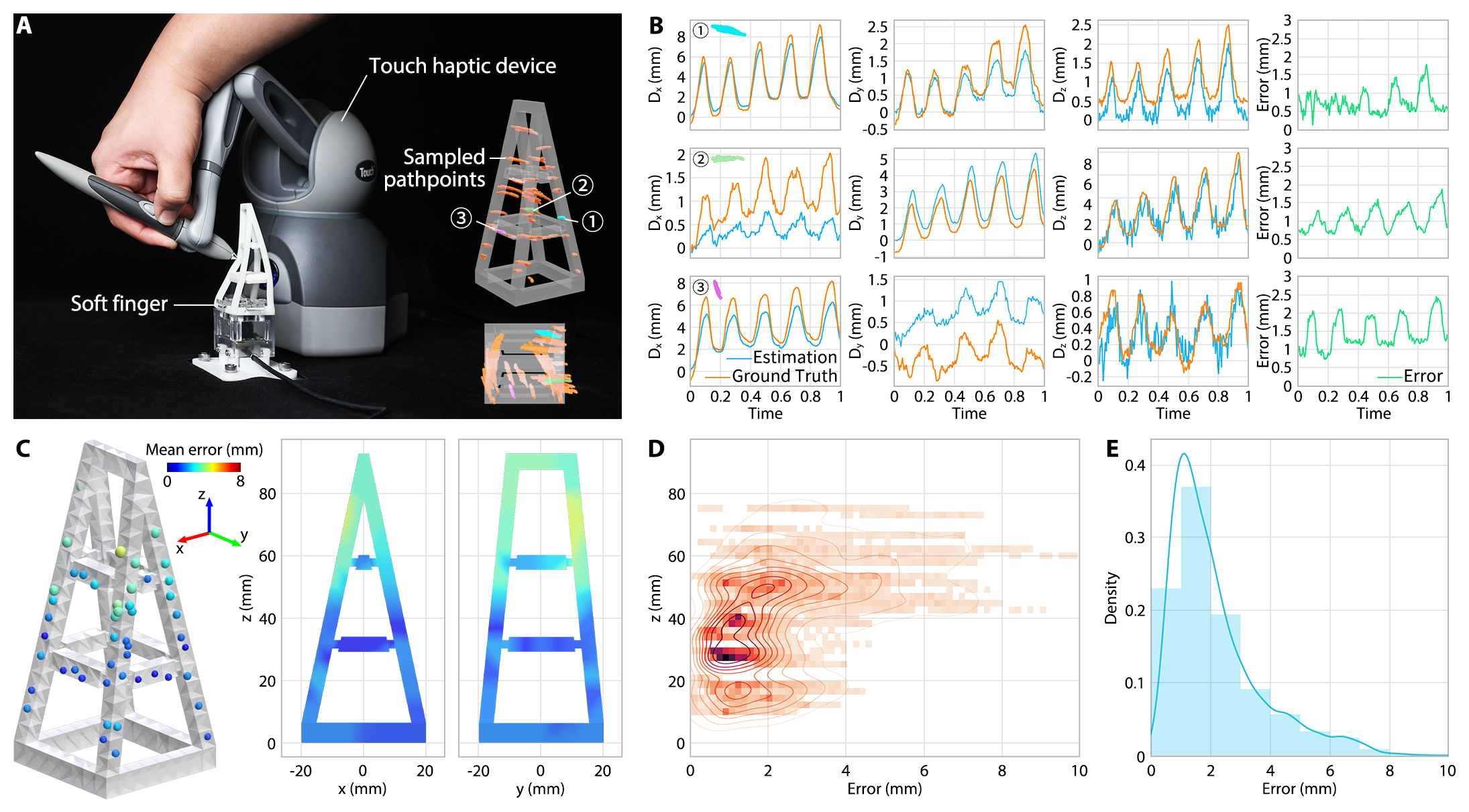}
        \caption{
        \textbf{Estimated deformation field of the soft finger using the proprioceptive state estimation method.} 
        (A) The Touch haptic device is used to make contact with the soft finger at different locations while simultaneously recording the ground-truth positions and the reconstructed positions of contact points. 
        (B) Three sampled pushing trajectories of the pen-nib and corresponding measurements from the proprioceptive state estimation method. Total Errors are reported in the last column. The pen-nib of the touch haptic device is pushed forward and backward five times at each location.
        (C) The fifty testing locations sampled are spread over half of the side of the soft finger. The mean error norm map is interpolated using the values of the fifty sampled contact locations.
        (D) The distribution of the total errors along the height (Z-axis) of the soft finger. 
        (E) The distribution of the total errors of sampled contact points.
        }
        \label{fig:Result-VBPropSE-Haptic}
    \end{figure*}

    We designed another validation experiment using the pen-nib's position of a haptic device (Touch by 3D Systems, Inc.) as ground truth measurement. As shown in Fig. \ref{fig:Result-VBPropSE-Haptic}A, an operator holding the pen-nib initiated contact at a random point on the soft robotic finger by pushing it five times. Fifty points were sampled, spreading over half of the soft robotic finger with recorded pen-nib position and the corresponding point of contact on the estimated deformation in the mesh model. 
    
    Similar to the calibration process when using the motion capture system, we solve the barycentric coordinates in Eq. \eqref{barycentric} using the initial contact position of pen-nib and the undeformed vertex position of the tetrahedron nearest to the contact point. Since there is no slipping between the contact point and the pen-nib, recording the pushing position of the pen-nib for a randomly selected point is equivalent to collecting the ground truth deformation field of the soft finger evaluated at that point. Figure \ref{fig:Result-VBPropSE-Haptic}B shows three selected pushing trajectories and the corresponding errors between estimation and ground truth. The pushing duration lasts around ten seconds for each location and is rescaled to 1 in the plot. The data is recorded at 20 Hz. Due to the variations among the pushing trajectories among the three locations, the errors are slightly different, but all lie within a 2.5 mm range.

    The haptic device measurements cover an extensive portion of the soft robotic finger, revealing further details regarding the spatial distribution of the estimation errors. We visualize the mean errors of deformation estimation evaluated at the fifty randomly selected contact locations in Fig. \ref{fig:Result-VBPropSE-Haptic}C. We interpolated two side views of continuous error distribution for the soft robotic finger with errors of all sampled locations using a Gaussian-kernel-based nearest-neighbor method \cite{Lehmann1999SurveyInterpolation}. 
    
    Contact locations near the observed AMH constraint are expected to exhibit fewer errors due to penalized computation near this region during deformation optimization. We plot the error distribution of all sampled locations along the $Z$ axis in Fig. \ref{fig:Result-VBPropSE-Haptic}D. Contact locations with a similar height to the AMH constraint exhibit a smaller and more concentrated error distribution. Figure \ref{fig:Result-VBPropSE-Haptic}E shows the error histogram of the overall experiment records, where the median of estimated error for the whole-body deformation is 1.96 mm, corresponding to 2.1\% of the finger's length. (See Movie S2 in the Supplementary Materials for a video demonstration.)

%%%%%%%%%%%%%%%%%%%%%%%%%%%%%%%%%
\subsection{On Amphibious Tactile Sensing for PropSE}
\label{sec:Result-Underwater}
%%%%%%%%%%%%%%%%%%%%%%%%%%%%%%%%%

    Here, we further investigate our proposed method in amphibious tactile sensing through three experiments in lab conditions. We begin by benchmarking our proposed VBTS method at controlled turbidity underwater. Then, we present a touch-based object shape reconstruction task to demonstrate the application of our proposed solution for amphibious tactile sensing. Finally, we present a full-system demonstration by attaching our robotic finger to the gripper of an underwater Remotely Operated Vehicle (ROV) for underwater grasping in a water tank, which we plan to implement further in the field test soon.

\subsubsection{Benchmarking VBTS Underwater against Turbidity}
\label{sec:Result-Underwater-LabTurbidity}

    Our proposed rigidity-aware AMH method effectively transforms the visual perception process for deformable shape reconstruction into a marker-based pose recognition problem. Therefore, the benchmarking of our vision-based tactile sensing solution underwater is directly determined by successfully recognizing the fiducial marker poses used in our system under different turbidity conditions. Turbidity is an optical characteristic that measures the clarity of a water body and is reported in Nephelometric Turbidity Units (NTU) \cite{Kitchener2017AReview}. It influences the visibility of optical cameras for underwater inspection, inducing light attenuation effects caused by the suspended particles \cite{Lee2022AutonomousUnderwater}. As one of the critical indicators for characterizing water quality, there have been rich studies on the turbidity of large water bodies worldwide. For example, previous research \cite{Li2021SpatialVariation} shows that the Yangtze River's turbidity is measured between 1.71 and 154 NTU.

    \begin{figure*}[htbp]
        \centering
        \includegraphics[width=\textwidth]{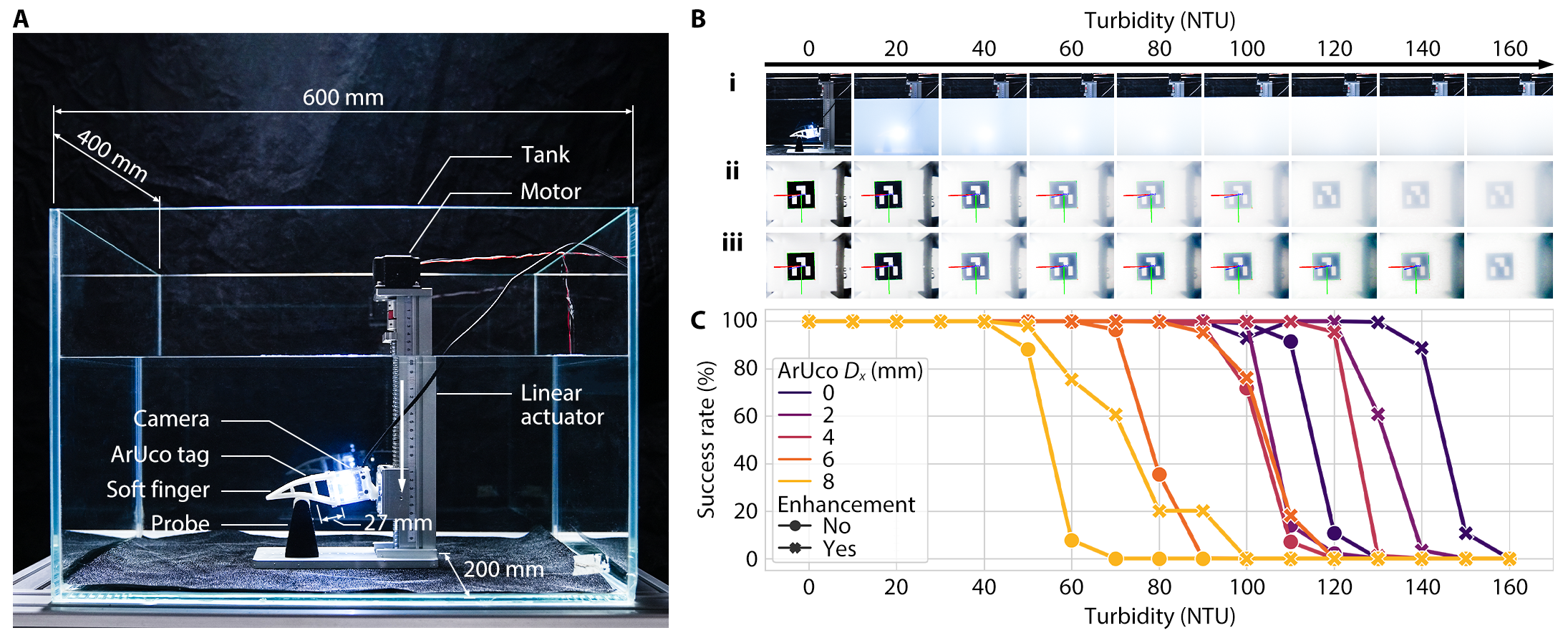}
        \caption{
        \textbf{Benchmarking results in different turbidity conditions underwater in a lab tank.}
        (A) The experiment was set up in a room with controlled ambient lighting of 3,000 lumens placed atop the tank (not shown in this picture).
        (B) Images taken by adding condensed standard turbidity liquid to increase the water turbidity from 0 to 160 NTU, including i) experiment pictures taken by an external camera at the same angle as (A); ii) raw images captured by the in-finger vision overlayed with triad coordinates to indicate successful pose recognition; and iii) digitally enhanced images overlayed with triad coordinates to indicate successful pose recognition. 
        (C) Results on the pose recognition success rate of the ArUco marker from the in-finger vision under increasing tank turbidity when pushing the soft robotic finger at different target displacements, with or without image enhancement.
        }
        \label{fig:Result-Underwater-LabTurbidity}
    \end{figure*}
    
    We investigated the robustness of our proposed VBTS solution in different water clarity conditions by mixing condensed standard turbidity liquid with clear water to reach different turbidity ratings. Figure \ref{fig:Result-Underwater-LabTurbidity}A shows the experiment setup. Our proposed soft robotic finger is installed on a linear actuator in a tank filled with 56 liters of clear water. A probe is fixed under the soft robotic finger, inducing contact-based whole-body deformation when the finger is commanded to move downward. The tank is placed in a room with controlled ambient lighting of 3,000 lumens placed atop the tank. We controlled the linear actuator for each turbidity condition so that the finger moved downward along the $x$ axis. This enabled us to record the ArUco image streams when fixed 0, 2, 4, 6, and 8 mm displacements in $D_x$ are reached. For example, the three images shown in the first column of Fig. \ref{fig:Result-Underwater-LabTurbidity}B are i) the experiment scenario taken at the same angle as Fig. \ref{fig:Result-Underwater-LabTurbidity}A when the turbidity is zero (before adding condensed standard turbidity liquid), ii) a sample of the raw image captured by our soft robotic finger's in-finger camera, and iii) image enhancement based on the image shown in ii), respectively. The water tank's clarity is modified by adding specific portions of condensed standard turbidity liquid to reach different turbidity ratings at 10 NTU per step (images for 20 NTU per step increase are shown in Fig. \ref{fig:Result-Underwater-LabTurbidity}B for the ease of visualization), increasing from 0 to 160 NTU, covering the Yangtze River's turbidity range. 

    For each of the $D_x$ positions, we recorded 1,000 images using our soft robotic finger's in-finger camera to obtain the pose recognition success rate (\%) under each turbidity rating, before or after image enhancement, reported in Fig. \ref{fig:Result-Underwater-LabTurbidity}C. The results reported in Fig. \ref{fig:Result-Underwater-LabTurbidity}C involve data aggregated from 85,000 raw images (1,000 images per NTU step per ArUco position $\times$ 17 NTU steps $\times$ 5 ArUco positions) from in-finger vision for ArUco pose recognition, which is doubled after image enhancement, resulting a total of 170K images. 
    
    In our experiment, for the turbidity range between 0 and 40 NTU, the raw images captured by our in-finger vision achieved a 100\% success rate in ArUco pose recognition. At 50 NTU turbidity, the first sign of failed marker pose recognition was observed when the most considerable deformation was induced at $8$ mm of $D_x$. Our experiment shows that this issue can be alleviated using simple image enhancement techniques to regain a 100\% marker pose recognition success rate. However, the marker pose recognition performance under large-scale whole-body deformation quickly deteriorated when the turbidity reached 60 NTU and eventually became unusable at 70 NTU. Image enhancement could effectively increase the upper bound to 100 NTU to reach an utterly unusable marker pose recognition in large-scale whole-body deformation. However, for small or medium whole-body deformations measured by $D_x \leq 6$ mm, our system remains functional until around 100 NTU in turbidity, where simple image enhancement techniques help for a balanced consideration between algorithmic cost, engineering complexity, and system performance. 

    \begin{figure*}[htbp]
        \centering
        \includegraphics[width=\textwidth]{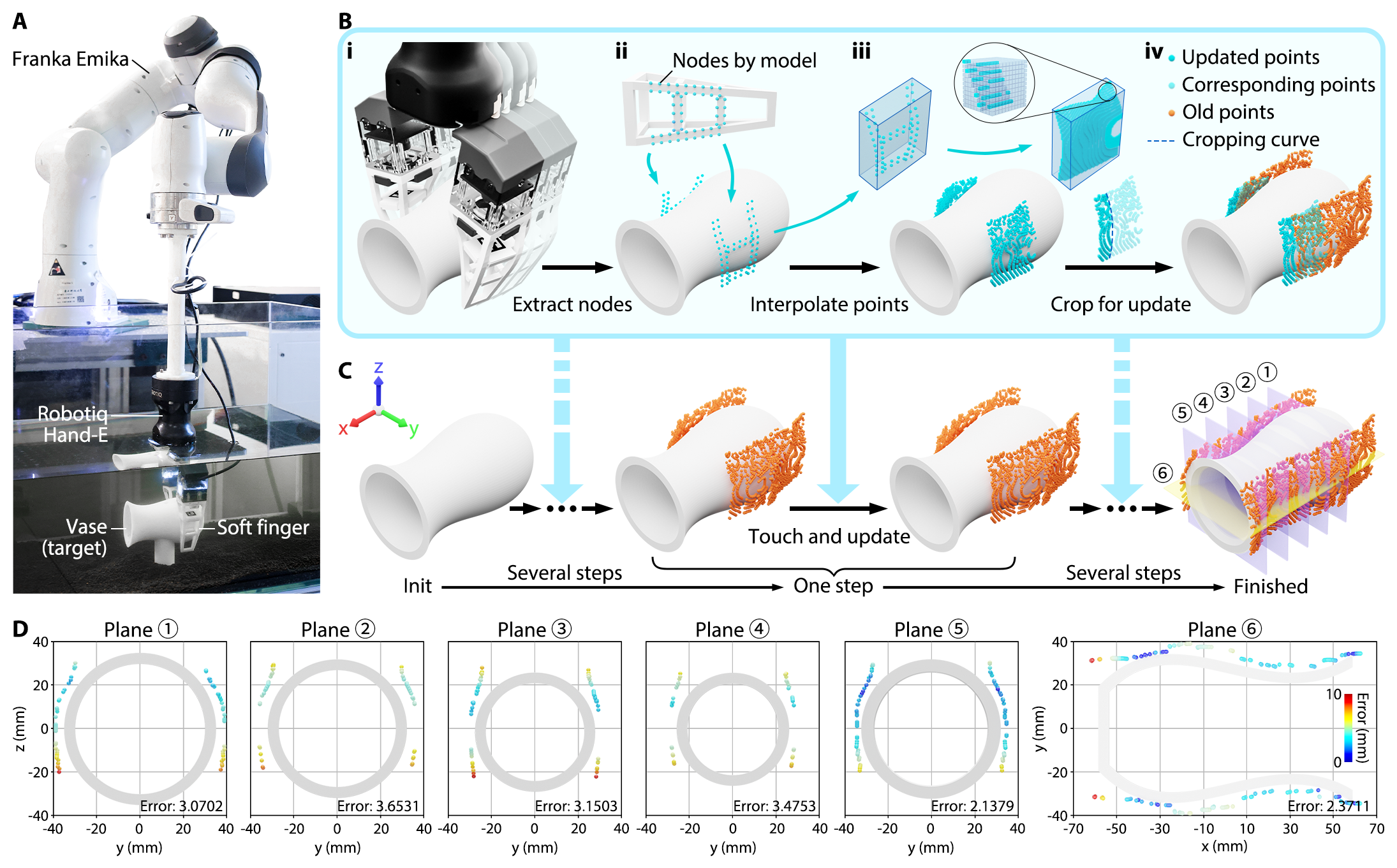}
        \caption{
        \textbf{Underwater shape estimation of a vase using proprioceptive state estimation of the soft finger.}
        (A) This is the experimental setup for underwater shape estimation. A Robotiq Hand-E gripper, installed with two proprioceptive soft fingers and an extension link, is mounted on a Franka Emika Panda robot arm. The gripper is programmed to perform a series of actions periodically, including gripping, releasing, and moving along the x-axis for a fixed distance. At the same time, a vase is fixed at the bottom of the tank in the lab. 
        (B) Contact surface patch prediction using Gaussian process implicit surface (GPIS) with the soft finger. 
        (C) Experiment pipeline for underwater shape estimation of a vase. 
        (D) Evaluation of the reconstructed vase shape on some cutting sectional planes, measured in Chamfer Distance.
        }
        \label{fig:Result-Underwater-LabShape}
    \end{figure*}
    
    For turbidity above 100 NTU, simple image enhancement provides limited contributions to our system. Our experiment shows that when the turbidity reached 160 NTU, our in-finger system failed to recognize any ArUco pose underwater, even after image enhancement. Since blurry images of the marker remain visible in the captured images, we can 1) use more advanced image processing algorithms, 2) use better imaging hardware, 3) apply stronger ambient lighting, and 4) redesign the marker pattern specifically for underwater usage to systematically increase the upper bound of the turbidity rating for marker-based posed estimation in contact-based amphibious grasping using vision-based tactile sensing methods. Results obtained from this experiment provide a general understanding of the potential regions for amphibious grasping characterized by turbidity with possible solutions to improve further.
    
\subsubsection{Underwater Exteroceptive Estimation of Object Shape} 
\label{sec:Result-Underwater-LabShape}

    In this experiment, we apply our soft robotic finger with in-finger vision to a contact-based shape reconstruction task to demonstrate our solution's capabilities in underwater exteroceptive estimation. Shown in Fig. \ref{fig:Result-Underwater-LabShape}A is the experimental setup conducted in the lab condition using the same water tank as the one used in the previous experiment. In this case, we used a parallel two-finger gripper (HandE from Robotiq, Inc.) attached to the wrist flange of a robotic manipulator (Franka Emika) through a 3D-printed cylindrical rod for an extended range of motion. Our soft robotic fingers are attached to each fingertip of the gripper through a customized adapter fabricated by 3D printing. Our previous work extensively tested this IP67 gripper's underwater servoing capabilities for reactive grasping during temporary submergence under the water \cite{Guo2024AutoencodingA}. In this study, we use the same gripper for underwater object shape estimation in a lab tank. One can always replace the Hand-E gripper with a professional underwater gripper for more intensive underwater usage in the field.

    With the gripper submerged underwater, the system is programmed to sequentially execute a series of actions, including gripping and releasing the object and moving along a prescribed direction for a fixed distance to acquire underwater object shape information, as shown in Fig. \ref{fig:Result-Underwater-LabShape}B(i). By mounting the target object at the bottom of the tank, we assume that 1) the object's pose is fixed and calibrated with the gripper and 2) passive object shape exploration is considered for object coverage. The inference of the GPIS model is computationally intractable for the large $N$ measurement that accrues from high-dimensional tactile measurements. Instead of predicting the whole object surface by accumulating all the collected data, we only query a local GPIS model approximated using current observed contact data in a local focus area and build the surface incrementally, as shown in Figs. \ref{fig:Result-Underwater-LabShape}B(ii)\&(iii).

    \paragraph{Local GPIS Model Inference} 
    
    A training set containing contact interface points $\mathbf{x}_{c}$ and corresponding augmented control points are collected each time a grasping action is performed. Before querying the local GPIS model in the interested area, hyper-parameters $\sigma^{2}_{f}$ and $l$ associated to Eq. \eqref{SE_COV} are optimized first using the standard training method for Gaussian processes, i.e., maximizing the marginal likelihood. Then, we evaluate the local GP on voxel grid points at a resolution of 0.2 mm in the interested area and keep those points with zero mean of Eq. \eqref{pred_mean} as estimated points on the surface patch of the object. 
    
    \paragraph{Local Patches Concatenation} 
    
    After calibrating the object pose to the gripper, we programmed the grasping system to follow a pre-defined path for object shape exploration. As is shown in Fig. \ref{fig:Result-Underwater-LabShape}B(iv), each time after GPIS query in the local 3D region, a global registration action is performed by transforming these local iso-surface points into the global space. Leveraging the continuous nature of the pre-defined exploration path, a simple surface concatenation strategy is used, where only the points of the estimated surface patch corresponding to moving distance are kept, and points of overlapping intervals belonging to the latest estimated surface patch are rejected. As is shown in Fig. \ref{fig:Result-Underwater-LabShape}C, after initialization of the relative pose between the gripper and the object, the shape of the object is continuously reconstructed using the described passive exploration strategy. 
    
    \paragraph{Object Shape Estimation Evaluation} 
    
    In Fig. \ref{fig:Result-Underwater-LabShape}D, we present our method on actual data collected during the underwater tactile exploration experiment. The shape estimates at each cutting sectional plane are compared concerning the ground truth using the Chamfer Distance (CD) \cite{Thayananthan2003ShapeContext}, a commonly-used shape similarity metric. We chose five vertical cutting planes and one horizontal sectional plane for reconstructed object surface evaluation. For each cutting plane, a calibration error exists between the vase and the Hand-E gripper, leading to the expected gap between the reconstructed and ground truth points. In addition to the systematic error, we have observed a slight decrease in the CD metric values between planes 1 and 5 compared to planes 2, 3, and 4, which could be attributed to the limitations of the soft finger in adapting to small objects with significant curvature. On the other hand, by employing tactile exploration actions with a relatively large contact area on the soft finger's surface, the shape estimation of objects similar in size to the vase can be accomplished more efficiently, typically within 8-12 touches. The 3D-printed vase has dimensions of approximately 80 mm by 80 mm by 140 mm. (See Movie S3 in the Supplementary Materials for a video demonstration.)

\subsubsection{Vision-based Tactile Grasping with an Underwater ROV}
\label{sec:Result-Underwater-LabDemo}

    % Here, we provide a full-system demonstration by using our vision-based soft robotic fingers to replace the existing fingers of a robotic gripper on a high-end underwater Remotely Operated Vehicle (ROV). This underwater ROV (FIFISH PRO V6 PLUS by QYSEA\footnote{https://www.qysea.com/}) is an enterprise-grade ROV for advanced underwater inspection. It features 6,000 lumens onboard lighting with 4K video recording capabilities, capable of diving up to 150 meters deep. The model we tested also includes a single-DOF robotic gripper, which can be modified by replacing the default fingers using customized adaptors. 
    Here, we provide a full-system demonstration by using our vision-based soft robotic fingers on a underwater Remotely Operated Vehicle (ROV, FIFISH PRO V6 PLUS by QYSEA\footnote{https://www.qysea.com/}). It includes a single-DOF robotic gripper, which can be modified using the proposed soft fingers with customized adaptors. 

    The experiment results reported in Section \ref{sec:Result-Underwater-LabTurbidity} already benchmark our system's promising capabilities for real-time underwater tactile sensing. As shown in Fig. \ref{fig:Result-Underwater-LabTurbidity}B, the water at 20 NTU or above is already very challenging to clearly observe from a third-personal perspective. Experiments underwater would require additional cost to prepare a second underwater ROV to record videos when the water is clear enough. However, as analyzed above, our in-finger vision could perform nicely at a much higher NTU range. Therefore, in this section, we only conducted this experiment in a lab tank to demonstrate our system's integration with an existing underwater ROV system during an underwater task. 

    \begin{figure}[htbp]
        \centering
        \includegraphics[width=0.8\textwidth]{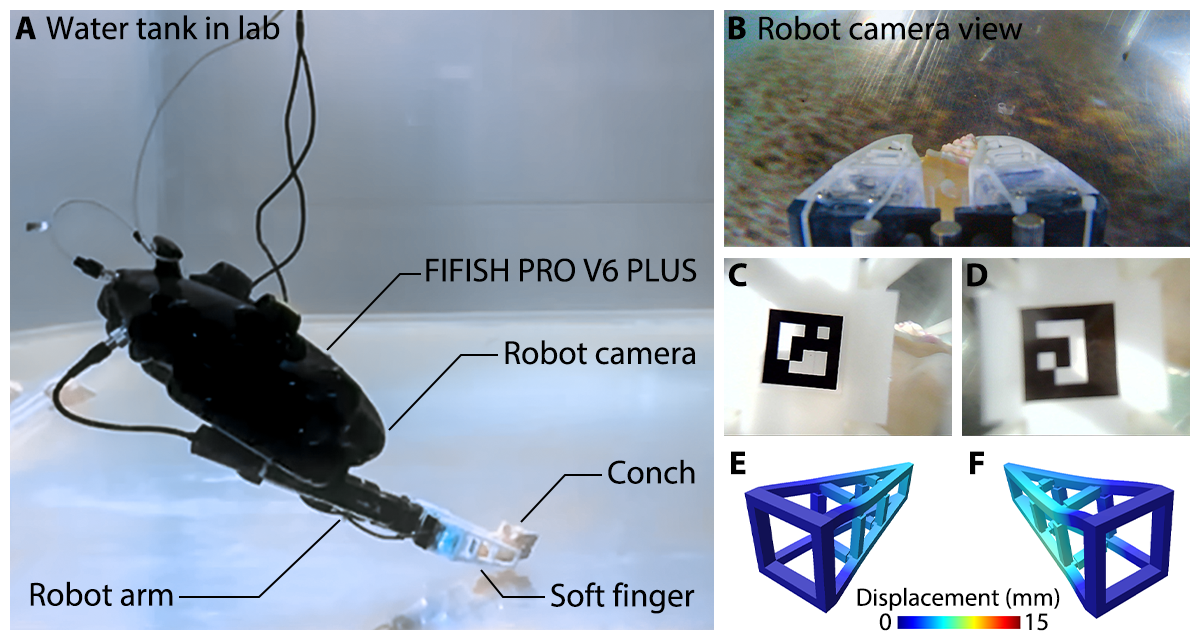}
        \caption{
        \textbf{Demonstration of our soft robotic finger with in-finger vision for tactile sensing underwater.}
        (A) Key components involved in the test. 
        (B) Screenshot of a 4K image captured by the underwater ROV's onboard camera when our fingers are holding a conch after successfully grasping.
        (C) \& (D) A screenshot of the images captured by an in-finger vision camera in the left and right fingers while holding the conch.
        (E) \& (F) Whole-body deformation reconstruction for both fingers based on the images captured by the in-finger vision cameras, respectively. 
        }
        \label{fig:Result-Underwater-LabDemo}
    \end{figure}

    Shown in Fig. \ref{fig:Result-Underwater-LabDemo}A is a brief overview of the system and the scene. Our fingers are attached to the underwater ROV's gripper through 3D-printed adaptors to replace the default rigid fingers. Our design conveniently introduced omni-directional adaptation capability to the gripper's existing functionality with added capabilities in real-time tactile sensing underwater. Shown in Fig. \ref{fig:Result-Underwater-LabDemo}B is a screenshot of the image taken by the ROV's onboard camera, recording 4K videos in real-time. In this experiment, both soft robotic fingers are installed with in-finger vision, capturing images shown in Figs. \ref{fig:Result-Underwater-LabDemo}C\&D. Using these in-finger images, we can use the methods proposed in this work to achieve real-time reconstruction of contact events on our soft robotic finger in Figs. \ref{fig:Result-Underwater-LabDemo}E\&F, while performing grasping tasks underwater.

    See Movie S4 in the Supplementary Materials for a video demonstration. Besides the capabilities demonstrated in this paper, we also identified an interesting observation during the experiment, adding to the benefits of having soft robotic fingers for underwater ROVs compared to the traditional rigid ones. While performing grasping underwater, the target objects are usually at the bottom. It is challenging for the underwater ROV to approach the target object smoothly and slowly, even in the lab tank with no water disturbances, which is also highly related to the pilot skills. Our soft fingers offer an added layer of production when the fingers collide with the bottom or other obstacles underwater, providing impact absorption for the underwater ROV while providing capable grasping and tactile sensing capabilities. If the original rigid fingers were installed, sudden impacts would occur when a collision happens, causing damage to the robot, the finger and gripper, and the underwater environment. 

%%%%%%%%%%%%%%%%%%%%%%%%%%%%%%%%%
\section{Discussion}
\label{sec:Discuss}
%%%%%%%%%%%%%%%%%%%%%%%%%%%%%%%%%

\subsection{Encoding Large-Scale, Whole-Body Deformation by Tracking a Single Visual Representation}

    This study presents a model-based representation by tracking a single visual feature to achieve high-performing reconstruction of large-scale, whole-body deformation for proprioceptive state estimation. We introduced rigidity-aware Aggregated Multi-Handle constraints during the modeling process. This problem is usually characterized by infinite degrees of freedom (DOFs) via a single visual feature in a 6D pose. As a result, we effectively reduced the dimensionality in representing soft, large-scale, whole-body deformation. Our method shows 40 to 700 times faster run-time than commercial software such as Abaqus at different resolutions while exhibiting superior accuracy in deformation reconstruction. Our method also shows 1 to 2 times faster than the widely adopted As-Rigid-As-Possible (ARAP) algorithm. It should be noted that it remains theoretically unsolved to provide a model-based explicit proof regarding this problem, requiring further research in future works. However, our study shows promising capabilities of this approach toward a high-performing solution with real-time reconstruction efficiency and accuracy that can be used for tactile robotics. 

\subsection{Rigid-Soft Interactive Representation in Tactile Robotics}

    The guiding principle behind our solution is a physical representation process shared by many existing solutions in Vision-Based Tactile Sensing (VBTS) technologies. Robotics usually interprets the physical world as an object-centric environment, which can be modeled as rigid bodies, soft bodies, or realistic bodies depending on predefined assumptions. A critical task in robotics is to provide a structured, digitalized representation of the unstructured, physical interactions so that the robotic system can make reliable action plans. The various designs of the soft medium in VBTS generally function as a \textit{physical filter} to transform unstructured, object-centric properties from the external environment into a constrained problem space within the finger towards a refined representation. In this study, we propose a rigid-soft interactive representation using a rigid body (the marker plate) attached to the soft body (the adaptive finger) during contact-based interactions (filled with realistic bodies with various material stiffness). This process is similar to the mass-point model in physics, which provides a succinate placeholder for deriving various physical properties without losing generality in the mathematical formulation. Further development following such representation principle may give researchers a novel perspective to model robotic dynamics as a tactile network of rigid-soft interactive representations, as demonstrated by results reported in this study. 
    
\subsection{Vision-Based Multi-Modal Tactile Sensing for Robotics}

    In this study, we focus our investigation in VBTS on deformation reconstruction only, which can further implement tactile sensing of other perceptual modalities, as demonstrated in our previous work. For example, our recent work \cite{Liu2024ProprioceptiveLearning} achieved state-of-the-art performance in 6D force-and-torque (FT) estimation using a similar design, where a fiducial marker is also attached inside the finger to provide a convenient representation. Combining both methods will achieve a Vision-Based Multi-Modal Tactile Sensing system in our soft robotic finger design, simultaneously providing high-performing tactile sensing in 6D FT and continuous whole-body deformation reconstruction. This will address a significant challenge in robot learning from demonstration \cite{Ravichandar2020AnnualReview, Chi2023DiffusionPolicy, Wu2024VisionBased}. Recent research \cite{Wan2023SeeThruFinger} also shows the possibility of achieving object detection in the external environment using the in-finger vision with a markerless design by implementing the in-painting technique. Our research provides a comprehensive demonstration regarding the robotic potentials of VBTS technology in fundamental theory and engineering applications, contributing to tactile robotics as a promising direction for future research \cite{Haddadin2018TactileRobots}. 

\subsection{Vision-based Tactile Sensing for Amphibious Robotics}

    Another novelty of this study is the application of VBTS in amphibious robotics. Our study presents comprehensive results and demonstrations in benchmarking performances, shape reconstruction tasks, and system integration with an underwater remotely operated vehicle. Many VBTS solutions require a closed chamber for the miniature camera to implement the photometric principle for tactile sensing, which may become challenging or even unrealistic for a direct application underwater. It should be noted that even after filling the closed chamber with a highly transparent resin to seal the camera, the layer of soft material used on the contact surface needs a depth-dependent calibration that is unrealistic to perform underwater. Furthermore, the soft material, such as silicon gel, will become brittle as the water depth increases \cite{Li2021SelfPowered}. Our previous work already showcased the engineering benefits of our soft robotic finger design, which can be used to reliably estimate 6D FT from on-land to underwater scenarios \cite{Guo2024AutoencodingA}. In this work, we further demonstrate the applications of VBTS in high-performing shape reconstruction through our soft robotic finger design for amphibious applications. Our soft finger's metamaterial network leverages structural adaptation by design instead of being solely dependent on the material softness. This significantly reduces the fluidic pressure on our finger's adaptive behavior. Further discussion of this topic is outside the scope of this study, which we will address in an upcoming work with more details.
    
%%%%%%%%%%%%%%%%%%%%%%%%%%%%%%%%%
\section{Conclusion, Limitations, and Future Work}
\label{sec:Conclude}
%%%%%%%%%%%%%%%%%%%%%%%%%%%%%%%%%

    In conclusion, this study presents a novel Vision-Based Tactile Sensing approach for Proprioceptive State Estimation with a focus on amphibious applications. Utilizing a Soft Polyhedral Network structure coupled with marker-based in-finger vision, our method achieves real-time, high-fidelity tactile sensing that accommodates omni-directional adaptations. The introduction of a model-based approach with rigidity-aware Aggregated Multi-Handle constraints enables effective optimization of the soft robotic finger’s deformation. Furthermore, restructuring our proposed approach as an implicit surface model demonstrates superior shape reconstruction and touch-point estimation performance compared to existing solutions. Experimental validations affirm its efficacy in large-scale reconstruction, turbidity benchmarking, and tactile grasping on an underwater Remotely Operated Vehicle, thereby highlighting the potential of tactile robotics for advanced amphibious applications.

    However, the study has several limitations. Manufacturing inconsistencies inherent to soft robots can impact the accuracy of our method, and algorithmic parameters require precise calibration through physical experiments. Additionally, using a rigid plate for boundary condition acquisition slightly hampers the finger's compliance, affecting the contact-based conformation between the object and our finger. The object surface estimation pipeline is also sensitive to contact geometry, restricting its use to local surface patches with smooth curvature changes.

    Future research aims to optimize the system for versatile tactile grasping and expand its integration into robotic grippers for diverse on-land and underwater applications. The vision-based proprioception method holds the potential for developing advanced robotic necks for underwater humanoids with precise state estimation driven by parallel mechanisms or pneumatic actuation. These advancements will pave the way for the broader application and utility of vision-based tactile sensing technologies in robotic systems operating in complex environments.

%%%%%%%%%%%%%%%%%%%%%%%%%%%%%%%%%
\section*{Supplementary Materials}
\label{sec:SupMat}
%%%%%%%%%%%%%%%%%%%%%%%%%%%%%%%%%

    \textbf{Movie S1. Evaluating Proprioceptive State Estimation using Motion Capture System.} This movie demonstrates the four contact configurations tested using the test platform and presents the error measurement protocol utilized in Section \ref{sec:Result-VBPropSE-MoCap}.
    
    \textbf{Movie S2. Estimating State when Deformed by Touch Haptic Device.} This movie showcases the experimental procedures described in Section \ref{sec:Result-VBPropSE-Haptic}. These involve measuring the position discrepancy between the pen-nib and the nearest node on the soft finger, representing the estimated deformation field error sampled at the corresponding location.
    
    \textbf{Movie S3. Shape Sensing of a Vase Underwater.} This movie features a demonstration of the experiment setup described in Section \ref{sec:Result-Underwater-LabShape}. It provides a comprehensive overview of the entire experimental process and showcases the results obtained from estimating the shape of an underwater vase using the soft finger.

    \textbf{Movie S4. Vision-based Tactile Grasping with an Underwater ROV} This movie demonstrates the experiment in Section \ref{sec:Result-Underwater-LabDemo}, where our soft robotic fingers with in-finger vision installed on the FIFISH Pro V6 Plus robot's gripper for tactile grasping underwater, providing omni-directional adaptation with real-time finger deformation reconstruction to perceive contact-events underwater. 

%%%%%%%%%%%%%%%%%%%%%%%%%%%%%%%%%
\section*{Appendix A: Abaqus Simulation}
%%%%%%%%%%%%%%%%%%%%%%%%%%%%%%%%%

    The nonlinear deformation computation of the volumetric soft finger is first carried out using Abaqus, an advanced finite element analysis (FEA) software.
    
\subsection*{Material Calibration}

\begin{figure}[ht]
        \centering
        \includegraphics[width=0.6\textwidth]{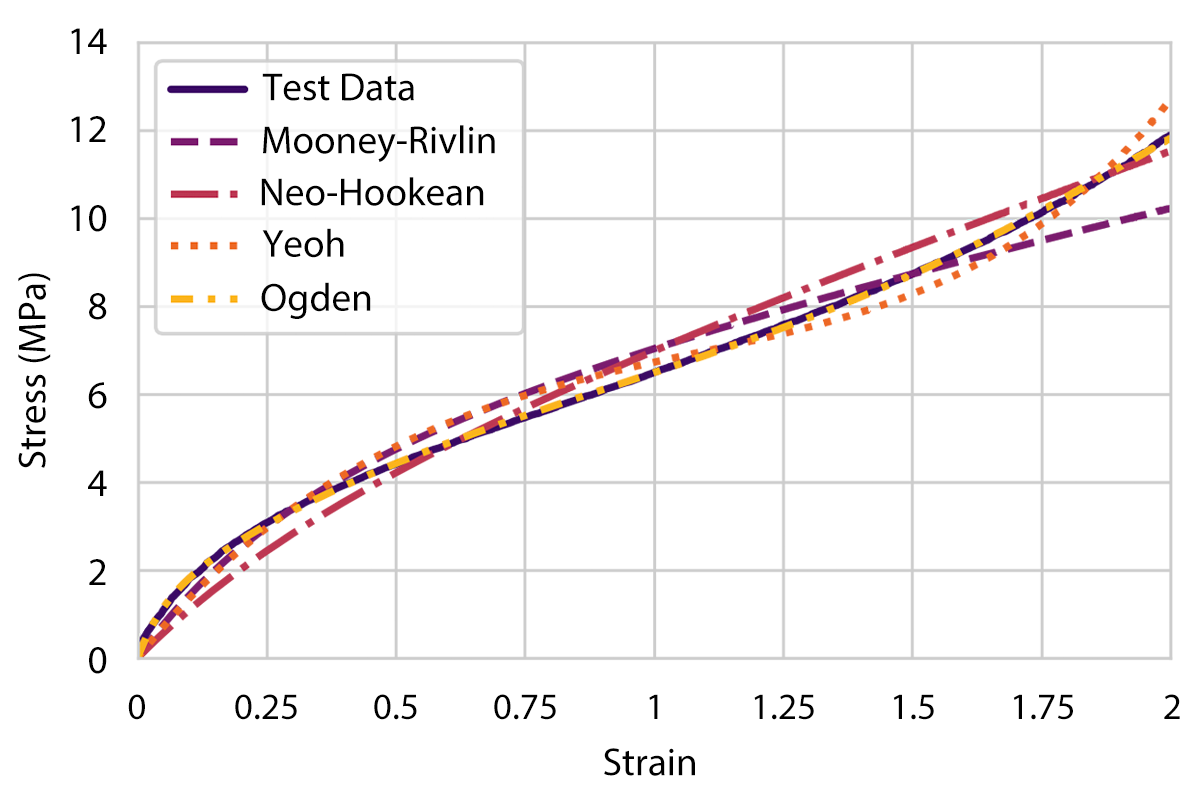}
        \caption{
        \textbf{Modeling hyperelastic behavior of the Hei-cast 8400 using uniaxial tension test data in Abaqus.}
        Several different strain energy potential models, including Mooney-Rivlin, Neo-Hookean, Yeoh, and Ogden (N=3), are selected to fit the test data, and Ogden is the best.
        }
        \label{fig:App-Abaqus-MaterialTesting}
    \end{figure}

    A uniaxial tension test is performed to accurately determine the material's mechanical properties in the soft finger. The 3rd-order Ogden hyperelastic model is found to be most appropriate for describing the material's mechanical behavior, as it shows a good match between the experimental stress-strain response and theoretical predictions, as illustrated in Fig. \ref{fig:App-Abaqus-MaterialTesting}.
        
\subsection*{Simulation Setup}

    In the simulation, one boundary condition, Encastre, secures the finger's bottom surface. Another boundary condition related to Displacement/Rotation is applied to the node set corresponding to the Aggregation Multi-Handle (AMH) constraints, which deforms the finger. The Abaqus analysis then provides the coordinates of all nodes in the finger mesh.

    To enhance the accuracy of simulations depicting the material behavior of Hei-cast 8400, six deviatoric coefficients are incorporated within the Ogden model. However, calibrating these coefficients requires a comprehensive set of mechanical experimental data. The calibration process is further complicated by the variations and inconsistencies typically encountered in the manufacturing processes of soft robotic components \cite{Esmail2020UsingThe}.

%%%%%%%%%%%%%%%%%%%%%%%%%%%%%%%%%
\section*{Appendix B: Implementation of the ARAP Method}
%%%%%%%%%%%%%%%%%%%%%%%%%%%%%%%%%

    The As-Rigid-As-Possible (ARAP) method is advantageous in interactive mesh deformation, animation, and 3D modeling, where the objective is to enable users to manipulate shapes while preserving local feature integrity.
    
\subsection*{Deformation Energy}
    
    Utilizing the same notation as in Eq. \eqref{eq:sd_element_energy}, the element-wise ARAP deformation energy function can be expressed as:
    \begin{equation}\label{eq:arap}
        \Psi_{ARAP}(\Phi_{t_{j}})=\Psi_{ARAP}{(\mathbf{A}_{t_{j}})}=||\mathbf{A}_{t_{j}}-\mathbf{R}||^2_\mathcal{F}
        ,
    \end{equation}
    where $\mathbf{R}$ represents the closest rotation to the deformation gradient $\mathbf{A}_{t{j}}\in {\mathbb{R}^{3\times{3}}}$ of the tetrahedron element $t_j$ in the Frobenius norm, defined as:
    \begin{equation}\label{eq:closest_rotation}
        \mathbf{R} = \mathop{\arg\min}\limits_{\mathbf{R}\in{SO(3)}} ||\mathbf{R} - \mathbf{A}_{t_{j}}||^2_\mathcal{F}
        .
    \end{equation}
    
    To calculate the deformed positions of the soft finger mesh nodes under the constraints set by AMHs using the ARAP energy model, the constrained geometric optimization problem in Eq. \eqref{eq:prime_fun} needs to be modified by replacing the symmetric Dirichlet energy Eq. \eqref{eq:sd_element_energy} with the ARAP energy in Eq. \eqref{eq:arap}.
    
\subsection*{Local/Global Optimization}
    
    Unlike the symmetric Dirichlet energy discussed in this article, the ARAP energy model does not permit a generic formulation of gradients and Hessians suitable for a Newton-type solver, such as Alg. \ref{pn}. To adapt the optimization process for deforming soft finger mesh nodes under constraints with the ARAP energy model, we utilized the local-global solver implemented using the \textit{libigl} library\footnote{https://libigl.github.io/}, a well-regarded C++ framework known for its efficiency in geometric computations.
    
    The local step concentrates on Eq. \eqref{eq:closest_rotation}, aiming to find the closest rotation from the deformation gradient of each mesh element. Subsequently, the global step minimizes the deviation of the deformation gradient from those rotations computed in the local step across all elements. The local and global steps are crucial in maintaining a balance between local shape preservation and the structure's overall integrity \cite{Liu2008ALocalGlobal}.
    
    In addressing the challenge of minimizing deformation energy under AMH constraints, this solver also transforms the constrained problem into an unconstrained one by applying a soft penalty method. We retain the same penalty parameter $\omega$ used in our implementation to ensure a fair comparison.

%%%%%%%%%%%%%%%%%%%%%%%%%%%%%%%%%
\section*{Appendix C: Parameter Selection for Algorithm \ref{pn}}
%%%%%%%%%%%%%%%%%%%%%%%%%%%%%%%%%

    As discussed in Section \ref{sec:Method-ModelPropSE-ShapeEst}, transforming the constrained geometry optimization problem Eq. \eqref{eq:prime_fun} into an unconstrained one Eq. \eqref{constr_fun} by establishing soft boundaries offers several advantages, especially considering the inherent variability in the accuracy of camera-observed constraints. Nonetheless, the potential deviation from constraints due to the soft boundary approach must be meticulously evaluated.
    
\begin{figure}[htbp]
        \centering
        \includegraphics[width=0.6\textwidth]{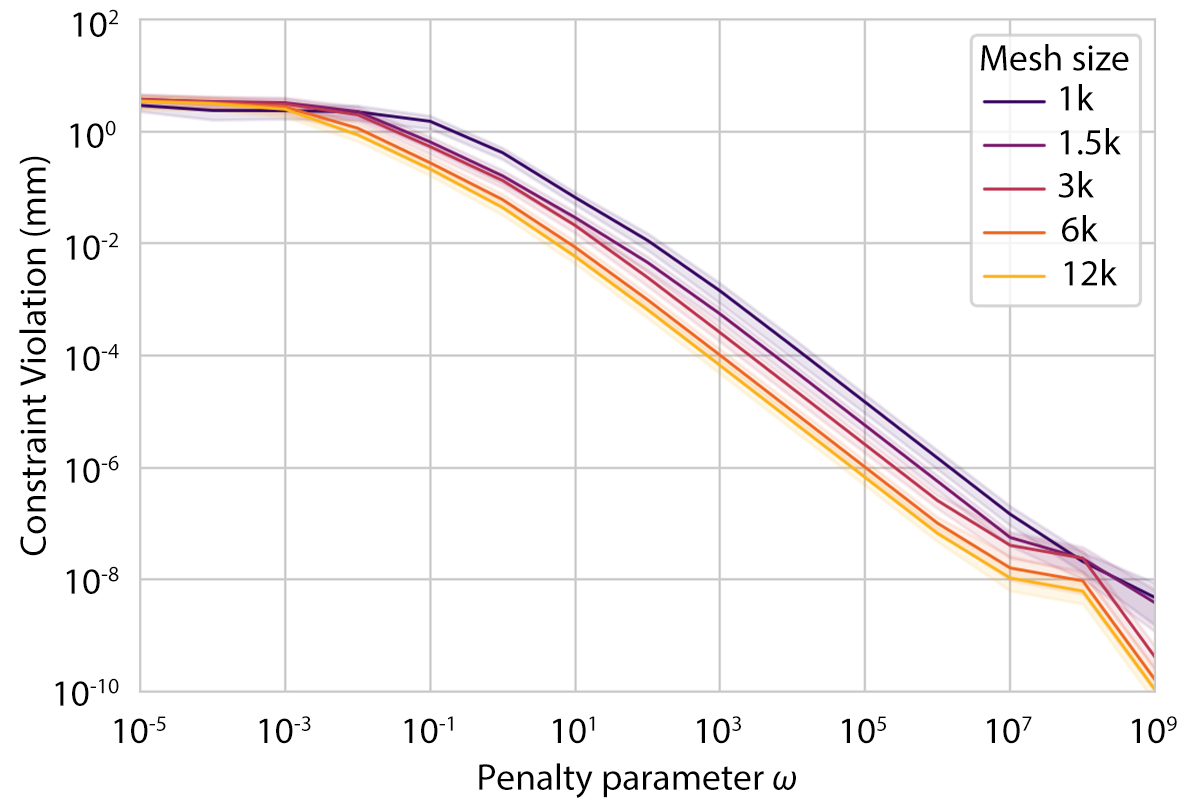}
        \caption{
        \textbf{Effect of soft penalty parameter $\omega$ on constraint violation.}}
        \label{fig:App-Alg1}
    \end{figure}

    Fig. \ref{fig:App-Alg1} depicts the relationship between constraint violations and different mesh sizes for a range of penalty parameter $\omega$ values. As $\omega$ increases, there is a consistent reduction in constraint violations, highlighting the effectiveness of adjusting the soft penalty parameter to improve compliance with specified constraints.
    
\begin{figure}[htbp]
        \centering
        \includegraphics[width=0.6\textwidth]{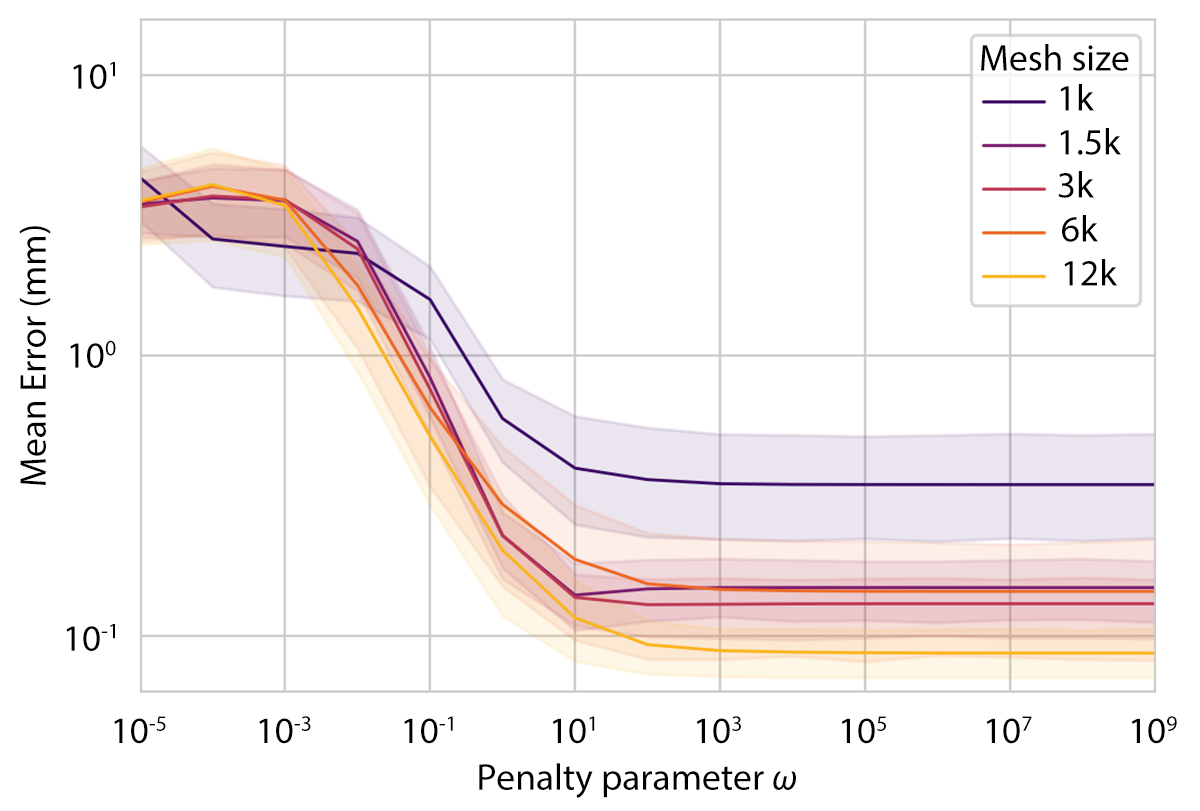}
        \caption{
        \textbf{Effect of soft penalty parameter $\omega$ on shape reconstruction accuracy.}}
        \label{fig:App-Alg1-MeanError}
    \end{figure}

    Moreover, Fig. \ref{fig:App-Alg1-MeanError} compares mean errors in geometric optimization against the ground truth established by Abaqus simulations. The data indicates that the mean error experiences less significant fluctuations with larger $\omega$ values.
    
    Based on the results of these empirical evaluations, we also consider the possibility of algorithmic improvements through strategic adjustments to the soft penalty parameter $\omega$ strategy. Specifically, we suggest that adopting an adaptive scheme, where $\omega$ is systematically increased with each iteration, or employing the augmented Lagrangian method to dynamically optimize $\omega$ for each iteration, could significantly enhance the performance of our algorithm. These approaches are expected to enable more accurate constraint handling and improve overall shape estimation accuracy by ensuring the optimal balance of $\omega$ in response to the evolving conditions of the optimization process.
    
    For this study, we have utilized a constant $\omega = 10^5$, motivated by our aim to establish a stable baseline to evaluate the core capabilities of our method. This fixed value was chosen to provide consistency across our experiments, facilitating a more straightforward assessment of the foundational performance of our shape estimation method.

%%%%%%%%%%%%%%%%%%%%%%%%%%%%%%%%%
\bibliographystyle{unsrt}
\bibliography{References}

\begin{thebibliography}{10}

\bibitem{Billard2019TrendsAnd}
Aude Billard and Danica Kragic.
\newblock \href{https://doi.org/10.1126/science.aat8414}{Trends and challenges
  in robot manipulation}.
\newblock {\em Science}, 364(6446):eaat8414, 2019.

\bibitem{Diaz2023MachineLearning}
Fernando D{\'\i}az~Ledezma and Sami Haddadin.
\newblock \href{https://doi.org/10.1126/scirobotics.adh0972}{Machine
  learning--driven self-discovery of the robot body morphology}.
\newblock {\em Science Robotics}, 8(85):eadh0972, 2023.

\bibitem{Van2018SoftOptoelectronic}
IM~Van~Meerbeek, CM~De~Sa, and RF~Shepherd.
\newblock \href{https://doi.org/10.1126/scirobotics.aau2489}{Soft
  optoelectronic sensory foams with proprioception}.
\newblock {\em Science Robotics}, 3(24):eaau2489, 2018.

\bibitem{Dahiya2009TactileSensing}
Ravinder~S Dahiya, Giorgio Metta, Maurizio Valle, and Giulio Sandini.
\newblock \href{https://doi.org/10.1109/TRO.2009.2033627}{Tactile
  Sensing—From Humans to Humanoids}.
\newblock {\em IEEE Transactions on Robotics}, 26(1):1--20, 2009.

\bibitem{Johansson2009CodingAnd}
Roland~S Johansson and J~Randall Flanagan.
\newblock \href{https://doi.org/10.1038/nrn2621}{Coding and Use of Tactile
  Signals from the Fingertips in Object Manipulation Tasks}.
\newblock {\em Nature Reviews Neuroscience}, 10(5):345--359, 2009.

\bibitem{Sundaram2019LearningThe}
Subramanian Sundaram, Petr Kellnhofer, Yunzhu Li, Jun-Yan Zhu, Antonio
  Torralba, and Wojciech Matusik.
\newblock \href{https://doi.org/10.1038/s41586-019-1234-z}{Learning the
  Signatures of the Human Grasp Using a Scalable Tactile Glove}.
\newblock {\em Nature}, 569(7758):698--702, 2019.

\bibitem{Chun2021AnArtificial}
Sungwoo Chun, Jong-Seok Kim, Yongsang Yoo, Youngin Choi, Sung~Jun Jung, Dongpyo
  Jang, Gwangyeob Lee, Kang-Il Song, Kum~Seok Nam, Inchan Youn, Donghee Son,
  Changhyun Pang, Yong Jeong, Hachul Jung, Young-Jin Kim, Byong-Deok Choi,
  Jaehun Kim, Sung-Phil Kim, Wanjun Park, and Seongjun Park.
\newblock \href{https://doi.org/10.1038/s41928-021-00585-x}{An Artificial
  Neural Tactile Sensing System}.
\newblock {\em Nature Electronics}, 4(6):429--438, 2021.

\bibitem{Shih2020ElectronicSkins}
Benjamin Shih, Dylan Shah, Jinxing Li, Thomas~G Thuruthel, Yong-Lae Park,
  Fumiya Iida, Zhenan Bao, Rebecca Kramer-Bottiglio, and Michael~T Tolley.
\newblock \href{https://doi.org/10.1126/scirobotics.aaz9239}{Electronic Skins
  and Machine Learning for Intelligent Soft Robots}.
\newblock {\em Science Robotics}, 5(41):eaaz9239, 2020.

\bibitem{Dahiya2013DirectionsToward}
Ravinder~S. Dahiya, Philipp Mittendorfer, Maurizio Valle, Gordon Cheng, and
  Vladimir~J. Lumelsky.
\newblock \href{https://doi.org/10.1109/JSEN.2013.2279056}{Directions Toward
  Effective Utilization of Tactile Skin: A Review}.
\newblock {\em IEEE Sensors Journal}, 13(11):4121--4138, 2013.

\bibitem{OpenAI2020LearningDexterous}
OpenAI:~Marcin Andrychowicz, Bowen Baker, Maciek Chociej, Rafal Jozefowicz, Bob
  McGrew, Jakub Pachocki, Arthur Petron, Matthias Plappert, Glenn Powell, Alex
  Ray, et~al.
\newblock \href{https://doi.org/10.1177/0278364919887447}{Learning Dexterous
  In-Hand Manipulation}.
\newblock {\em The International Journal of Robotics Research}, 39(1):3--20,
  2020.

\bibitem{Kappassov2015TactileSensing}
Zhanat Kappassov, Juan-Antonio Corrales, and V{\'e}ronique Perdereau.
\newblock \href{https://doi.org/10.1016/j.robot.2015.07.015}{Tactile Sensing in
  Dexterous Robot Hands}.
\newblock {\em Robotics and Autonomous Systems}, 74:195--220, 2015.

\bibitem{Yousef2011TactileSensing}
Hanna Yousef, Mehdi Boukallel, and Kaspar Althoefer.
\newblock \href{https://doi.org/10.1016/j.sna.2011.02.038}{Tactile Sensing for
  Dexterous In-Hand Manipulation in Robotics — A Review}.
\newblock {\em Sensors and Actuators A: Physical}, 167(2):171--187, 2011.

\bibitem{Wan2022VisualLearning}
Fang Wan, Xiaobo Liu, Ning Guo, Xudong Han, Feng Tian, and Chaoyang Song.
\newblock \href{https://proceedings.mlr.press/v164/wan22a}{Visual Learning
  Towards Soft Robot Force Control using a 3D Metamaterial with Differential
  Stiffness}.
\newblock In {\em Conference on Robot Learning}, pages 1269--1278. PMLR, 2022.

\bibitem{Liu2024ProprioceptiveLearning}
Xiaobo Liu, Xudong Han, Wei Hong, Fang Wan, and Chaoyang Song.
\newblock \href{https://doi.org/10.1177/02783649241238765}{Proprioceptive
  Learning with Soft Polyhedral Networks}.
\newblock {\em The International Journal of Robotics Research}, 0(0):1--20,
  2024.

\bibitem{Kim2020HeterogeneousSensing}
Taekyoung Kim, Sudong Lee, Taehwa Hong, Gyowook Shin, Taehwan Kim, and Yong-Lae
  Park.
\newblock \href{https://doi.org/10.1126/scirobotics.abc6878}{Heterogeneous
  Sensing in a Multifunctional Soft Sensor for Human-Robot Interfaces}.
\newblock {\em Science Robotics}, 5(49):eabc6878, 2020.

\bibitem{Faure2012SOFA}
Fran{\c{c}}ois Faure, Christian Duriez, Herv{\'e} Delingette, J{\'e}r{\'e}mie
  Allard, Benjamin Gilles, St{\'e}phanie Marchesseau, Hugo Talbot, Hadrien
  Courtecuisse, Guillaume Bousquet, Igor Peterlik, and St{\'e}phane Cotin.
\newblock {\em \href{https://doi.org/10.1007/8415_2012_125}{SOFA: A Multi-Model
  Framework for Interactive Physical Simulation}}, pages 283--321.
\newblock Springer Berlin Heidelberg, Berlin, Heidelberg, 2012.

\bibitem{Stuart2017OceanOne}
Hannah Stuart, Shiquan Wang, Oussama Khatib, and Mark~R Cutkosky.
\newblock \href{https://doi.org/10.1177/0278364917694723}{The Ocean One Hands:
  An Adaptive Design for Robust Marine Manipulation}.
\newblock {\em The International Journal of Robotics Research}, 36(2):150--166,
  2017.

\bibitem{Suresh2021TactileSLAM}
Sudharshan Suresh, Maria Bauza, Kuan-Ting Yu, Joshua~G. Mangelson, Alberto
  Rodriguez, and Michael Kaess.
\newblock \href{https://doi.org/10.1109/ICRA48506.2021.9562060}{Tactile SLAM:
  Real-Time Inference of Shape and Pose from Planar Pushing}.
\newblock In {\em IEEE International Conference on Robotics and Automation
  (ICRA)}, pages 11322--11328, 2021.

\bibitem{Mazzeo2022MarineRobotics}
Angela Mazzeo, Jacopo Aguzzi, Marcello Calisti, Simonepietro Canese, Fabrizio
  Vecchi, Sergio Stefanni, and Marco Controzzi.
\newblock \href{https://doi.org/10.3390/s22020648}{Marine Robotics for Deep-Sea
  Specimen Collection: A Systematic Review of Underwater Grippers}.
\newblock {\em Sensors}, 22(2):648, 2022.

\bibitem{Meerbeek2018SoftOptoelectronic}
I.~M.~Van Meerbeek, C.~M.~De Sa, and R.~F. Shepherd.
\newblock \href{https://doi.org/10.1126/scirobotics.aau2489}{Soft
  Optoelectronic Sensory Foams with Proprioception}.
\newblock {\em Science Robotics}, 3(24):eaau2489, 2018.

\bibitem{Li2020AReview}
Qiang Li, Oliver Kroemer, Zhe Su, Filipe~Fernandes Veiga, Mohsen Kaboli, and
  Helge~Joachim Ritter.
\newblock \href{https://doi.org/10.1109/TRO.2020.3003230}{A Review of Tactile
  Information: Perception and Action Through Touch}.
\newblock {\em IEEE Transactions on Robotics}, 36(6):1619--1634, 2020.

\bibitem{DeMaria2012ForceTactile}
Giuseppe De~Maria, Ciro Natale, and Salvatore Pirozzi.
\newblock \href{https://doi.org/10.1016/j.sna.2011.12.042}{Force/Tactile Sensor
  for Robotic Applications}.
\newblock {\em Sensors and Actuators A: Physical}, 175:60--72, 2012.

\bibitem{Magrini2014EstimationOf}
Emanuele Magrini, Fabrizio Flacco, and Alessandro De~Luca.
\newblock \href{https://doi.org/10.1109/IROS.2014.6942848}{Estimation of
  Contact Forces Using a Virtual Force Sensor}.
\newblock In {\em IEEE/RSJ International Conference on Intelligent Robots and
  Systems (IROS)}, pages 2126--2133. IEEE, 2014.

\bibitem{Holladay2021PlanningFor}
Rachel Holladay, Tom{\'a}s Lozano-P{\'e}rez, and Alberto Rodriguez.
\newblock \href{https://doi.org/10.1109/ICRA48506.2021.9561233}{Planning for
  Multi-Stage Forceful Manipulation}.
\newblock In {\em IEEE International Conference on Robotics and Automation
  (ICRA)}, pages 6556--6562. IEEE, 2021.

\bibitem{Lin2020ContactSurface}
Hsiu-Chin Lin and Michael Mistry.
\newblock \href{https://doi.org/10.1109/ICRA40945.2020.9196816}{Contact Surface
  Estimation via Haptic Perception}.
\newblock In {\em IEEE International Conference on Robotics and Automation
  (ICRA)}, pages 5087--5093. IEEE, 2020.

\bibitem{Haddadin2018TactileRobots}
Sami Haddadin, Lars Johannsmeier, and Fernando~D{\'\i}az Ledezma.
\newblock \href{https://doi.org/10.1109/JPROC.2018.2879870}{Tactile Robots as a
  Central Embodiment of the Tactile Internet}.
\newblock {\em Proceedings of the IEEE}, 107(2):471--487, 2018.

\bibitem{Yan2022TactileSuperResolution}
Youcan Yan, Yajing Shen, Chaoyang Song, and Jia Pan.
\newblock \href{https://doi.org/10.1109/LRA.2022.3141449}{Tactile
  Super-Resolution Model for Soft Magnetic Skin}.
\newblock {\em IEEE Robotics and Automation Letters}, 7(2):2589--2596, 2022.

\bibitem{Wu2018ASkinInspired}
Yuanzhao Wu, Yiwei Liu, Youlin Zhou, Qikui Man, Chao Hu, Waqas Asghar, Fali Li,
  Zhe Yu, Jie Shang, Gang Liu, et~al.
\newblock \href{https://doi.org/10.1126/scirobotics.aat0429}{A Skin-Inspired
  Tactile Sensor for Smart Prosthetics}.
\newblock {\em Science Robotics}, 3(22):eaat0429, 2018.

\bibitem{Liu2022PrintedSynaptic}
Fengyuan Liu, Sweety Deswal, Adamos Christou, Mahdieh Shojaei~Baghini, Radu
  Chirila, Dhayalan Shakthivel, Moupali Chakraborty, and Ravinder Dahiya.
\newblock \href{https://doi.org/10.1126/scirobotics.abl7286}{Printed Synaptic
  Transistor-Based Electronic Skin for Robots to Feel and Learn}.
\newblock {\em Science Robotics}, 7(67):eabl7286, 2022.

\bibitem{Yan2021SoftMagnetic}
Youcan Yan, Zhe Hu, Zhengbao Yang, Wenzhen Yuan, Chaoyang Song, Jia Pan, and
  Yajing Shen.
\newblock \href{https://doi.org/10.1126/scirobotics.abc8801}{Soft Magnetic Skin
  for Super-Resolution Tactile Sensing with Force Self-Decoupling}.
\newblock {\em Science Robotics}, 6(51):eabc8801, 2021.

\bibitem{Yuan2017Gelsight}
Wenzhen Yuan, Siyuan Dong, and Edward~H Adelson.
\newblock \href{https://doi.org/10.3390/s17122762}{GelSight: High-Resolution
  Robot Tactile Sensors for Estimating Geometry and Force}.
\newblock {\em Sensors}, 17(12):2762, 2017.

\bibitem{WardCherrier2018TacTip}
Benjamin Ward-Cherrier, Nicholas Pestell, Luke Cramphorn, Benjamin Winstone,
  Maria~Elena Giannaccini, Jonathan Rossiter, and Nathan~F Lepora.
\newblock \href{https://doi.org/10.1089/soro.2017.0052}{The TacTip Family: Soft
  Optical Tactile Sensors with 3D-Printed Biomimetic Morphologies}.
\newblock {\em Soft Robotics}, 5(2):216--227, 2018.

\bibitem{Alspach2019SoftBubble}
Alex Alspach, Kunimatsu Hashimoto, Naveen Kuppuswamy, and Russ Tedrake.
\newblock \href{https://doi.org/10.1109/ROBOSOFT.2019.8722713}{Soft-Bubble: A
  Highly Compliant Dense Geometry Tactile Sensor for Robot Manipulation}.
\newblock In {\em IEEE International Conference on Soft Robotics (RoboSoft)},
  pages 597--604. IEEE, 2019.

\bibitem{Sun2022GuidingThe}
Huanbo Sun and Georg Martius.
\newblock \href{https://doi.org/10.1126/scirobotics.abm0608}{Guiding the Design
  of Superresolution Tactile Skins with Taxel Value Isolines Theory}.
\newblock {\em Science Robotics}, 7(63):eabm0608, 2022.

\bibitem{Trueeb2020TowardsVisionBased}
Camill Trueeb, Carmelo Sferrazza, and Raffaello D’Andrea.
\newblock \href{https://doi.org/10.1109/RoboSoft48309.2020.9116060}{Towards
  Vision-Based Robotic Skins: A Data-Driven, Multi-Camera Tactile Sensor}.
\newblock In {\em IEEE International Conference on Soft Robotics (RoboSoft)},
  pages 333--338. IEEE, 2020.

\bibitem{Sferrazza2019GroundTruth}
Carmelo Sferrazza, Adam Wahlsten, Camill Trueeb, and Raffaello D’Andrea.
\newblock \href{https://doi.org/10.1109/ACCESS.2019.2956882}{Ground Truth Force
  Distribution for Learning-Based Tactile Sensing: A Finite Element Approach}.
\newblock {\em IEEE Access}, 7:173438--173449, 2019.

\bibitem{Yamaguchi2019RecentProgress}
Akihiko Yamaguchi and Christopher~G Atkeson.
\newblock \href{https://doi.org/10.1080/01691864.2019.1632222}{Recent Progress
  in Tactile Sensing and Sensors for Robotic Manipulation: Can We Turn Tactile
  Sensing into Vision?}
\newblock {\em Advanced Robotics}, 33(14):661--673, 2019.

\bibitem{Armanini2023SoftRobots}
Costanza Armanini, Fr{\'e}d{\'e}ric Boyer, Anup~Teejo Mathew, Christian Duriez,
  and Federico Renda.
\newblock \href{https://doi.org/10.1109/TRO.2022.3231360}{Soft Robots Modeling:
  A Structured Overview}.
\newblock {\em IEEE Transactions on Robotics}, 39(3):1728--1748, 2023.

\bibitem{Yamaguchi2017ImplementingTactile}
Akihiko Yamaguchi and Christopher~G Atkeson.
\newblock \href{https://doi.org/10.1109/HUMANOIDS.2017.8246881}{Implementing
  Tactile Behaviors Using FingerVision}.
\newblock In {\em IEEE-RAS International Conference on Humanoid Robotics
  (Humanoids)}, pages 241--248. IEEE, 2017.

\bibitem{Kroeger2016FastOptical}
Till Kroeger, Radu Timofte, Dengxin Dai, and Luc Van~Gool.
\newblock \href{https://doi.org/10.1007/978-3-319-46493-0_29}{Fast Optical Flow
  Using Dense Inverse Search}.
\newblock In {\em European Conference on Computer Vision (ECCV)}. Springer,
  Cham, 2016.

\bibitem{Sferrazza2019DesignMotivation}
Carmelo Sferrazza and Raffaello D’Andrea.
\newblock \href{https://doi.org/10.3390/s19040928}{Design, Motivation and
  Evaluation of a Full-Resolution Optical Tactile Sensor}.
\newblock {\em Sensors}, 19(4):928, 2019.

\bibitem{Zhang2018VisionBasedSensing}
Zhongkai Zhang, J{\'e}r{\'e}mie Dequidt, and Christian Duriez.
\newblock \href{https://doi.org/10.1109/LRA.2018.2800781}{Vision-Based Sensing
  of External Forces Acting on Soft Robots Using Finite Element Method}.
\newblock {\em IEEE Robotics and Automation Letters}, 3(3):1529--1536, 2018.

\bibitem{Kaboli2019TactileBasedActive}
Mohsen Kaboli, Kunpeng Yao, Di~Feng, and Gordon Cheng.
\newblock \href{https://doi.org/10.1007/s10514-018-9707-8}{Tactile-Based Active
  Object Discrimination and Target Object Search in an Unknown Workspace}.
\newblock {\em Autonomous Robots}, 43:123--152, 2019.

\bibitem{Ilonen2014ThreeDimensionalObject}
Jarmo Ilonen, Jeannette Bohg, and Ville Kyrki.
\newblock \href{https://doi.org/10.1177/0278364913497816}{Three-Dimensional
  Object Reconstruction of Symmetric Objects by Fusing Visual and Tactile
  Sensing}.
\newblock {\em The International Journal of Robotics Research}, 33(2):321--341,
  2014.

\bibitem{Wang20183DShape}
Shaoxiong Wang, Jiajun Wu, Xingyuan Sun, Wenzhen Yuan, William~T Freeman,
  Joshua~B Tenenbaum, and Edward~H Adelson.
\newblock \href{https://doi.org/10.1109/IROS.2018.8593430}{3D Shape Perception
  from Monocular Vision, Touch, and Shape Priors}.
\newblock In {\em IEEE International Conference on Intelligent Robots and
  Systems (IROS)}, pages 1606--1613. IEEE, 2018.

\bibitem{Yasa2023AnOverview}
Oncay Yasa, Yasunori Toshimitsu, Mike~Y Michelis, Lewis~S Jones, Miriam
  Filippi, Thomas Buchner, and Robert~K Katzschmann.
\newblock \href{https://doi.org/10.1146/annurev-control-062322-100607}{An
  Overview of Soft Robotics}.
\newblock {\em Annual Review of Control, Robotics, and Autonomous Systems},
  6:1--29, 2023.

\bibitem{Baines2022MultiEnvironmentRobotic}
Robert Baines, Sree~Kalyan Patiballa, Joran Booth, Luis Ramirez, Thomas Sipple,
  Andonny Garcia, Frank Fish, and Rebecca Kramer-Bottiglio.
\newblock \href{https://doi.org/10.1038/s41586-022-05188-w}{Multi-Environment
  Robotic Transitions Through Adaptive Morphogenesis}.
\newblock {\em Nature}, 610(7931):283--289, 2022.

\bibitem{Rafeeq2021LocomotionStrategies}
Mohammed Rafeeq, Siti~Fauziah Toha, Salmiah Ahmad, and Mohd~Asyraf Razib.
\newblock \href{https://doi.org/10.1109/ACCESS.2021.3057406}{Locomotion
  Strategies for Amphibious Robots-A Review}.
\newblock {\em IEEE Access}, 9:26323--26342, 2021.

\bibitem{Yu2012OnA}
Junzhi Yu, Rui Ding, Qinghai Yang, Min Tan, Weibing Wang, and Jianwei Zhang.
\newblock \href{https://doi.org/10.1109/TMECH.2011.2132732}{On a Bio-inspired
  Amphibious Robot Capable of Multimodal Motion}.
\newblock {\em IEEE/ASME Transactions on Mechatronics}, 17(5):847--856, 2012.

\bibitem{Subad2021SoftRobotic}
Rafsan Al Shafatul~Islam Subad, Liam~B. Cross, and Kihan Park.
\newblock \href{https://doi.org/10.3390/applmech2020021}{Soft Robotic Hands and
  Tactile Sensors for Underwater Robotics}.
\newblock {\em Applied Mechanics}, 2(2):356--382, 2021.

\bibitem{Li2023AnAerial}
Lei Li, Wenbo Liu, Bocheng Tian, Peiyu Hu, Wenzhuo Gao, Yuchen Liu, Fuqiang
  Yang, Youning Duo, Hongru Cai, Yiyuan Zhang, Zhouhao Zhang, Zimo Li, and
  Li~Wen.
\newblock \href{https://doi.org/10.1002/aisy.202300381}{An Aerial--Aquatic
  Hitchhiking Robot with Remora-Inspired Tactile Sensors and Thrust Vectoring
  Units}.
\newblock {\em Advanced Intelligent Systems}, page 2300381 (Early View), 2023.

\bibitem{Aggarwal2015HapticObject}
Achint Aggarwal, Peter Kampmann, Johannes Lemburg, and Frank Kirchner.
\newblock \href{https://doi.org/10.1002/rob.21538}{Haptic Object Recognition in
  Underwater and Deep-sea Environments}.
\newblock {\em Journal of Field Robotics}, 32(1):167--185, 2015.

\bibitem{Yang2023DynamicCapture}
Shangkui Yang, Yongxiang Zhou, Ian~D. Walker, Chenghao Yang, David~T. Branson,
  Zhibin Song, Jian~Sheng Dai, and Rongjie Kang.
\newblock \href{https://doi.org/10.1109/TMECH.2022.3219108}{Dynamic Capture
  Using a Traplike Soft Gripper With Stiffness Anisotropy}.
\newblock {\em IEEE/ASME Transactions on Mechatronics}, 28(3):1337--1346, 2023.

\bibitem{Garrido2014AutomaticGeneration}
Sergio Garrido-Jurado, Rafael Mu{\~n}oz-Salinas, Francisco~Jos{\'e}
  Madrid-Cuevas, and Manuel~Jes{\'u}s Mar{\'\i}n-Jim{\'e}nez.
\newblock \href{https://doi.org/10.1016/j.patcog.2014.01.005}{Automatic
  Generation and Detection of Highly Reliable Fiducial Markers under
  Occlusion}.
\newblock {\em Pattern Recognition}, 47(6):2280--2292, 2014.

\bibitem{Lanczos1986TheVariational}
Cornelius Lanczos.
\newblock {\em \href{https://isbnsearch.org/isbn/9780486650678}{The Variational
  Principles of Mechanics (Dover Books on Physics, 4th Edition)}}.
\newblock Dover Publications, 1986.

\bibitem{Longva2020HigherOrderFinite}
Andreas Longva, Fabian L\"{o}schner, Tassilo Kugelstadt, Jos\'{e}~Antonio
  Fern\'{a}ndez-Fern\'{a}ndez, and Jan Bender.
\newblock \href{https://doi.org/10.1145/3414685.3417853}{Higher-Order Finite
  Elements for Embedded Simulation}.
\newblock {\em ACM Transactions on Graphics}, 39(6), 2020.

\bibitem{Aigerman2013Injective}
Noam Aigerman and Yaron Lipman.
\newblock \href{https://doi.org/10.1145/2461912.2461931}{Injective and Bounded
  Distortion Mappings in 3D}.
\newblock {\em ACM Transactions on Graphics}, 32(4):1--14, 2013.

\bibitem{Liu2008ALocalGlobal}
Ligang Liu, Lei Zhang, Yin Xu, Craig Gotsman, and Steven~J Gortler.
\newblock \href{https://doi.org/10.1111/j.1467-8659.2008.01290.x}{A
  Local/Global Approach to Mesh Parameterization}.
\newblock {\em Computer Graphics Forum}, 27(5):1495--1504, 2008.

\bibitem{Aigerman2015SeamlessSurface}
Noam Aigerman, Roi Poranne, and Yaron Lipman.
\newblock \href{https://doi.org/10.1145/2766921}{Seamless Surface Mappings}.
\newblock {\em ACM Transactions on Graphics}, 34(4), 2015.

\bibitem{Smith2015BijectiveParameterization}
Jason Smith and Scott Schaefer.
\newblock \href{https://doi.org/10.1145/2766947}{Bijective Parameterization
  with Free Boundaries}.
\newblock {\em ACM Transactions on Graphics}, 34(4), 2015.

\bibitem{Lai2009IntroductionTo}
W~Michael Lai, David Rubin, and Erhard Krempl.
\newblock {\em \href{https://isbnsearch.org/isbn/9780080509136}{Introduction to
  Continuum Mechanics (3rd Edition)}}.
\newblock Butterworth-Heinemann, 2014.

\bibitem{Rabinovich2017ScalableLocally}
Michael Rabinovich, Roi Poranne, Daniele Panozzo, and Olga Sorkine-Hornung.
\newblock \href{https://doi.org/10.1145/2983621}{Scalable Locally Injective
  Mappings}.
\newblock {\em ACM Transactions on Graphics}, 36(2), 2017.

\bibitem{Tretschk2023StateOf}
Edith Tretschk, Navami Kairanda, Mallikarjun BR, Rishabh Dabral, Adam
  Kortylewski, Bernhard Egger, Marc Habermann, Pascal Fua, Christian Theobalt,
  and Vladislav Golyanik.
\newblock \href{https://doi.org/10.1111/cgf.14774}{State of the Art in Dense
  Monocular Non-Rigid 3D Reconstruction}.
\newblock {\em Computer Graphics Forum}, 42(2):485--520, 2023.

\bibitem{GerardoCastro2015LaserRadarData}
Marcos~P Gerardo-Castro, Thierry Peynot, and Fabio Ramos.
\newblock \href{https://doi.org/10.13140/2.1.2702.6569}{Laser-Radar Data Fusion
  with Gaussian Process Implicit Surfaces}.
\newblock In {\em International Conference on Field and Service Robotics
  (FSR)}, pages 289--302. Springer, 2015.

\bibitem{Dragiev2011GaussianProcess}
Stanimir Dragiev, Marc Toussaint, and Michael Gienger.
\newblock \href{https://doi.org/10.1109/ICRA.2011.5980395}{Gaussian Process
  Implicit Surfaces for Shape Estimation and Grasping}.
\newblock In {\em IEEE International Conference on Robotics and Automation
  (ICRA)}, pages 2845--2850. IEEE, 2011.

\bibitem{Ottenhaus2016LocalImplicit}
Simon Ottenhaus, Martin Miller, David Schiebener, Nikolaus Vahrenkamp, and
  Tamim Asfour.
\newblock \href{https://doi.org/10.1109/HUMANOIDS.2016.7803372}{Local Implicit
  Surface Estimation for Haptic Exploration}.
\newblock In {\em IEEE International Conference on Humanoid Robots
  (Humanoids)}, pages 850--856. IEEE, 2016.

\bibitem{Rasmussen2005GaussianProcesses}
Carl~Edward Rasmussen and Christopher K.~I. Williams.
\newblock {\em \href{https://doi.org/10.7551/mitpress/3206.001.0001}{Gaussian
  Processes for Machine Learning}}.
\newblock The MIT Press, 2005.

\bibitem{Peltzer2020EigenAD}
Patrick Peltzer, Johannes Lotz, and Uwe Naumann.
\newblock \href{https://doi.org/10.1007/978-3-030-50371-0_51}{Eigen-AD:
  Algorithmic Differentiation of the Eigen Library}.
\newblock In {\em International Conference on Computational Science (ICCS)},
  page 690–704. Springer-Verlag, 2020.

\bibitem{Fang2020KinematicsOf}
Guoxin Fang, Christopher-Denny Matte, Rob~BN Scharff, Tsz-Ho Kwok, and
  Charlie~CL Wang.
\newblock \href{https://doi.org/10.1109/TRO.2020.2985583}{Kinematics of Soft
  Robots by Geometric Computing}.
\newblock {\em IEEE Transactions on Robotics}, 36(4):1272--1286, 2020.

\bibitem{Lehmann1999SurveyInterpolation}
Thomas~Martin Lehmann, Claudia Gonner, and Klaus Spitzer.
\newblock \href{https://doi.org/10.1109/42.816070}{Survey: Interpolation
  Methods in Medical Image Processing}.
\newblock {\em IEEE Transactions on Medical Imaging}, 18(11):1049--1075, 1999.

\bibitem{Kitchener2017AReview}
Ben~GB Kitchener, John Wainwright, and Anthony~J Parsons.
\newblock \href{https://doi.org/10.1177/030913331772654}{A Review of the
  Principles of Turbidity Measurement}.
\newblock {\em Progress in Physical Geography}, 41(5):620--642, 2017.

\bibitem{Lee2022AutonomousUnderwater}
Hoosang Lee, Daehyeon Jeong, Hongje Yu, and Jeha Ryu.
\newblock \href{https://doi.org/10.1007/s12555-021-0357-9}{Autonomous
  Underwater Vehicle Control for Fishnet Inspection in Turbid Water
  Environments}.
\newblock {\em International Journal of Control, Automation and Systems},
  20(10):3383--3392, 2022.

\bibitem{Li2021SpatialVariation}
Jianhong Li, Changchun Huang, Yong Zha, Chuan Wang, Nana Shang, and Weiyue Hao.
\newblock \href{https://doi.org/10.13227/j.hjkx.202103245}{Spatial Variation
  Characteristics and Remote Sensing Retrieval of Total Suspended Matter in
  Surface Water of the Yangtze River}.
\newblock {\em Environmental Science}, 42(12):5239--5249, 2021.

\bibitem{Guo2024AutoencodingA}
Ning Guo, Xudong Han, Xiaobo Liu, Shuqiao Zhong, Zhiyuan Zhou, Jian Lin,
  Jiansheng Dai, Fang Wan, and Chaoyang Song.
\newblock \href{https://doi.org/10.1002/aisy.202300382}{Autoencoding a Soft
  Touch to Learn Grasping from On-Land to Underwater}.
\newblock {\em Advanced Intelligent Systems}, 6(1):2300382, 2024.

\bibitem{Thayananthan2003ShapeContext}
A.~Thayananthan, B.~Stenger, P.H.S. Torr, and R.~Cipolla.
\newblock \href{https://doi.org/10.1109/CVPR.2003.1211346}{Shape Context and
  Chamfer Matching in Cluttered Scenes}.
\newblock In {\em IEEE Computer Society Conference on Computer Vision and
  Pattern Recognition (CVPR)}, 2003.

\bibitem{Ravichandar2020AnnualReview}
Harish Ravichandar, Athanasios~S. Polydoros, Sonia Chernova, and Aude Billard.
\newblock \href{https://doi.org/10.1146/annurev-control-100819-063206}{Recent
  Advances in Robot Learning from Demonstration}.
\newblock {\em Annual Review of Control, Robotics, and Autonomous Systems},
  3(Volume 3, 2020):297--330, 2020.

\bibitem{Chi2023DiffusionPolicy}
Cheng Chi, Siyuan Feng, Yilun Du, Zhenjia Xu, Eric Cousineau, Benjamin
  Burchfiel, and Shuran Song.
\newblock
  \href{https://roboticsconference.org/2023/program/papers/026/}{Diffusion
  Policy: Visuomotor Policy Learning via Action Diffusion}.
\newblock In {\em Proceedings of Robotics: Science and Systems (RSS)}, 2023.

\bibitem{Wu2024VisionBased}
Tianyu Wu, Yujian Dong, Xiaobo Liu, Xudong Han, Yang Xiao, Jinqi Wei, Fang Wan,
  and Chaoyang Song.
\newblock \href{https://doi.org/10.1016/j.matdes.2024.112629}{Vision-based
  Tactile Intelligence with Soft Robotic Metamaterial}.
\newblock {\em Materials \& Design}, 238:112629, 2024.

\bibitem{Wan2023SeeThruFinger}
Fang Wan and Chaoyang Song.
\newblock \href{https://doi.org/10.48550/arXiv.2312.09822}{SeeThruFinger: See
  and Grasp Anything with a Soft Touch}.
\newblock {\em arXiv:2312.09822 [cs.RO]}, 2023.

\bibitem{Li2021SelfPowered}
Guorui Li, Xiangping Chen, Fanghao Zhou, Yiming Liang, Youhua Xiao, Xunuo Cao,
  Zhen Zhang, Mingqi Zhang, Baosheng Wu, Shunyu Yin, Yi~Xu, Hongbo Fan, Zheng
  Chen, Wei Song, Wenjing Yang, Binbin Pan, Jiaoyi Hou, Weifeng Zou, Shunping
  He, Xuxu Yang, Guoyong Mao, Zheng Jia, Haofei Zhou, Tiefeng Li, Shaoxing Qu,
  Zhongbin Xu, Zhilong Huang, Yingwu Luo, Tao Xie, Jason Gu, Shiqiang Zhu, and
  Wei Yang.
\newblock \href{https://doi.org/10.1038/s41586-020-03153-z}{Self-powered soft
  robot in the Mariana Trench}.
\newblock {\em Nature}, 591(7848):66--71, 2021.

\bibitem{Esmail2020UsingThe}
Jihan~F Esmail, Mohammed~Z Mohamedmeki, and Awadh~E Ajeel.
\newblock \href{https://doi.org/10.1088/1757-899X/888/1/012065}{Using the
  Uniaxial Tension Test to Satisfy the Hyperelastic Material Simulation in
  ABAQUS}.
\newblock In {\em IOP Conference Series: Materials Science and Engineering},
  volume 888, page 012065. IOP Publishing, 2020.

\end{thebibliography}
%%%%%%%%%%%%%%%%%%%%%%%%%%%%%%%%%
\end{document}